\documentclass{article}

     \PassOptionsToPackage{numbers, compress}{natbib}

     \usepackage[final]{neurips_2025}

\usepackage[utf8]{inputenc} 
\usepackage[T1]{fontenc}    
\usepackage{hyperref}       
\usepackage{url}            
\usepackage{booktabs}       
\usepackage{amsfonts}       
\usepackage{nicefrac}       
\usepackage{microtype}      
\usepackage{xcolor}         
\usepackage{graphicx}
\usepackage{amsmath}
\usepackage{multirow}
\usepackage{wrapfig}

\title{GeGS-PCR: Effective and Robust 3D Point Cloud Registration with Two-Stage Color-Enhanced Geometric-3DGS Fusion}

\author{
	Jiayi Tian\textsuperscript{1}, Haiduo Huang\textsuperscript{1}, Tian Xia\textsuperscript{1}, Wenzhe Zhao\textsuperscript{1}, Pengju Ren\textsuperscript{1, }\thanks{Pengju Ren is the corresponding author}\\
	\textsuperscript{1} National Key Laboratory of Human-Machine Hybrid Augmented Intelligence, \\
	National Engineering Research Center of Visual Information and Applications, \\
	Institute of Artificial Intelligence and Robotics, Xi'an Jiaotong University, \\
	Xi'an, Shaanxi, China \\
	\texttt{\{tianreg, huanghd\}}@stu.xjtu.edu.cn \\
	\texttt{\{tian\_xia, wenzhe, pengjuren\}}@xjtu.edu.cn
}

\begin{document}

\maketitle

\begin{abstract}
  We address the challenge of point cloud registration using color information, where traditional methods relying solely on geometric features often struggle in low-overlap and incomplete scenarios. To overcome these limitations, we propose GeGS-PCR, a novel two-stage method that combines geometric, color, and Gaussian information for robust registration. Our approach incorporates a dedicated color encoder that enhances color features by extracting multi-level geometric and color data from the original point cloud. We introduce the \textbf{Ge}ometric-3D\textbf{GS} module, which encodes the local neighborhood information of colored superpoints to ensure a globally invariant geometric-color context. Leveraging LORA optimization, we maintain high performance while preserving the expressiveness of 3DGS. Additionally, fast differentiable rendering is utilized to refine the registration process,  leading to improved convergence. To further enhance performance, we propose a joint photometric loss that exploits both geometric and color features. This enables strong performance in challenging conditions with extremely low point cloud overlap. We validate our method by colorizing the Kitti dataset as ColorKitti and testing on both Color3DMatch and Color3DLoMatch datasets. Our method achieves state-of-the-art performance with \textit{Registration Recall} at 99.9\%, \textit{Relative Rotation Error} as low as 0.013, and \textit{Relative Translation Error} as low as 0.024, improving precision by at least a factor of 2.
\end{abstract}
\section{Introduction}
Fast and stable point cloud registration is a crucial technology in computer vision \cite{chen2023dreg} and embodied intelligence \cite{sautier2022image}, serving as the foundation for various practical applications, such as 3D scene reconstruction \cite{kerbl20233d}, semantic scene segmentation \cite{shafiullah2022clip}, and large-scale perception and mapping \cite{zhu2022nice}. In essence, point cloud registration involves aligning two overlapping 3D point clouds using a rigid transformation through a series of estimation steps. \\
Recent advancements in deep learning have accelerated the development of 3D point cloud representation \cite{yew2022regtr, ao2021spinnet} and differentiable optimization techniques \cite{choy2019fully, bai2020d3feat}. Previous works have focused on keypoints and correspondences, leveraging specialized neural networks to extract features from point clouds, and subsequently determining the rigid transformation using robust estimators like RANSAC \cite{deng2018ppfnet, huang2020feature}. Inspired by image matching, recent research has highlighted the significance of local neighborhood information \cite{sun2021loftr, zhou2021patch2pix}, by matching keypoints (superpoints) based on the detection of overlapping patches. This has led to advances in point cloud registration methods \cite{zhang20233d, qin2022geometric}, where downsampling is used to layer point clouds, and the Transformer architecture is employed to capture contextual information, adding informative constraints to the registration process. Furthermore, unsupervised correspondence-based point cloud registration methods that focus on optimizing point and Gaussian distribution correspondences \cite{fu2021robust, yuan2020deepgmr} have gained significant attention.

\begin{wrapfigure}{r}{0.65\textwidth} 
	\centering
	\includegraphics[width=\linewidth]{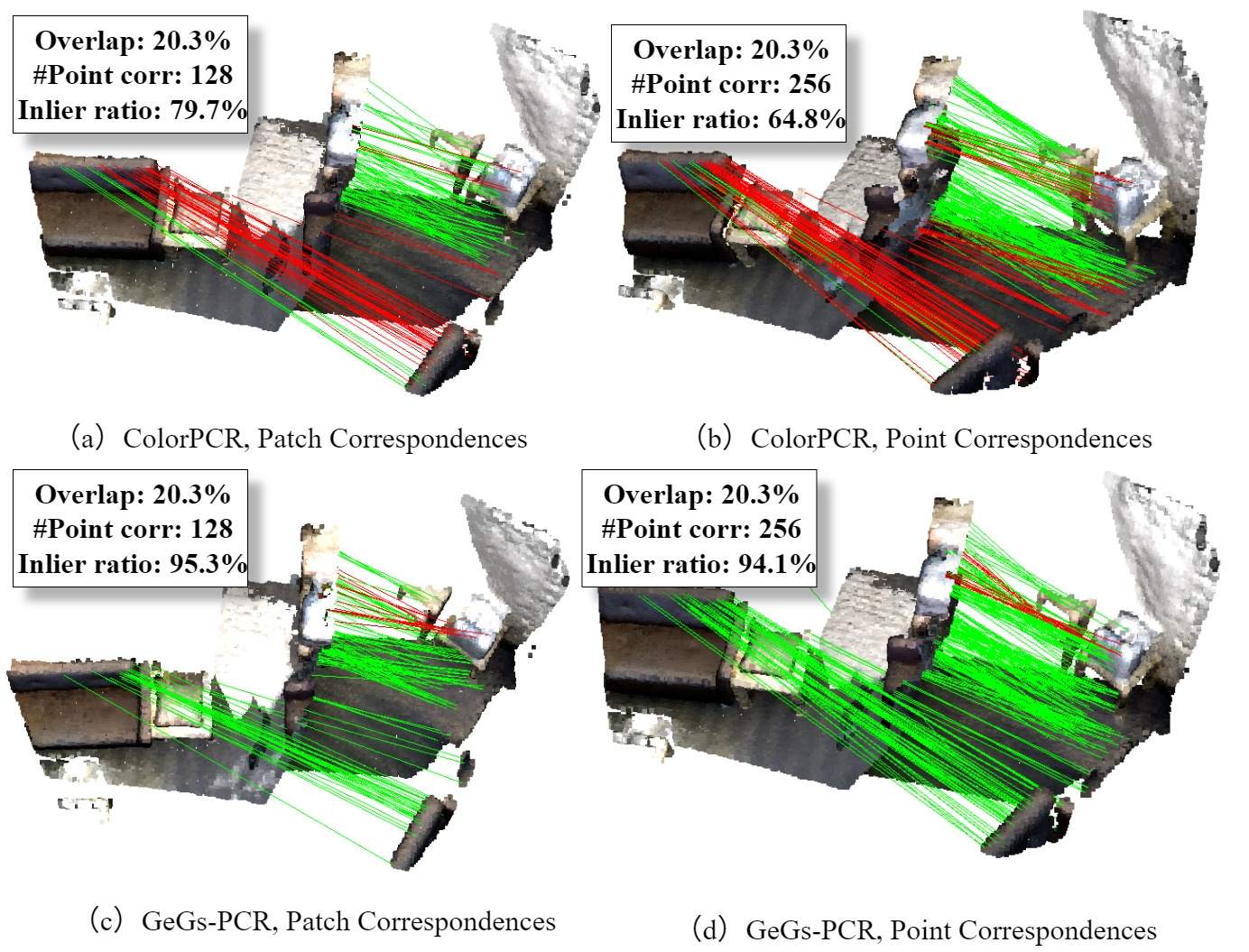}
	\caption{In scenarios with minimal overlap, incomplete geometric features, and subtle color variations, methods that simply add color features perform moderately, whereas GeGS-PCR successfully identifies the brown sofa.}
	\label{fig1}
\end{wrapfigure}
Despite rapid progress, point cloud registration remains challenging in real-world scenarios with low overlap between point clouds \cite{huang2020feature, choy2020deep}, where registration often fails. This highlights the need for novel methods. Recent breakthroughs in colored point cloud registration \cite{yu2023peal, mu2024colorpcr} have shown that integrating color information can reveal relationships that cannot be captured by geometric features alone, thereby improving registration performance. As shown in Fig.~\ref{fig1}, color information plays a crucial role in scenarios with low overlap and subtle geometric features. When color differences are not distinct, simply incorporating color information still fails to establish the correct correspondences. However, when subfigures (c) and (d) specifically analyze color features, the registration process can successfully match the brown sofa with its armrest. The Gaussian distribution captures global shape variations and significantly suppresses noise. Combining color and geometric information, along with considering the relationships between point clouds and their Gaussian distributions, is crucial for faster and more robust registration.\\
To address the challenges of point cloud registration in low-overlap real-world scenarios, we propose GeGS-PCR, a two-stage method that integrates Geometric-3DGS for colored point cloud registration. We designed a dedicated color encoder that enhances color features and extract multi-level geometric and color information from the original point cloud. The Geometric-3DGS module encodes local neighborhood information of colored superpoints, achieving globally invariant geographic color context. Using the parameterized multimodal local neighborhood information from 3DGS(geometric, Gaussian, and color information), we perform fast coarse registration. To reduce the computational complexity and parameter count introduced by 3DGS parameterization without sacrificing expressive power, we incorporate LORA optimization \cite{hu2022lora}. For better model convergence, we use 3DGS's fast differentiable rendering to refine the point cloud registration. Additionally, we introduce a joint photometric loss to improve the utilization of color information during the registration process. Through these modules, geometric and color data are tightly integrated, enabling GeGS-PCR to deliver strong performance even in challenging low-overlap scenarios.\\
Furthermore, due to the limited availability of publicly available colored datasets, currently only the publicly available COLOR3DMatch and Color3DLoMatch datasets \cite{mu2024colorpcr} are available. Therefore, to validate the generalization ability of the model, we colorized the Kitti dataset to create ColorKitti. Evaluations on these datasets demonstrate that GeGS-PCR can achieve fast and stable registration under low overlap, proving its advanced effectiveness.\\
The main contributions of this paper are as follows:
\begin{itemize}
\item We tightly combine color and geometric information to achieve point cloud registration. Specifically, we design a color encoder for feature extraction, constructing a globally invariant geometric-color representation.
\item We propose the Geometric-3DGS module to encode multimodal representations of superpoint neighborhood information. Using attention with 3DGS embeddings, we focus on global geometric distribution-color features and perform fast coarse registration by reducing computational complexity with LORA.
\item We introduce a joint photometric loss. By performing fast differentiable rendering on the Geometric-3DGS module and calculating photometric loss during the rendering process, we refine the registration of point clouds.
\item We colorize public point cloud datasets to generate ColorKitti. Experimental validation shows that GeGS-PCR performs excellently even in scenarios with extremely low overlap.
\end{itemize}
\section{Related Work}
\subsection{Correspondence-Based Methods}
Existing correspondence-based methods can be roughly divided into point-to-point and point-to-distribution registration approaches. Point-to-point methods (e.g., ICP) aim to estimate transformations through point coordinates or feature extraction \cite{121791}, \cite{9009829}. Using robust pose estimators (such as RANSAC or other RANSAC-free methods \cite{zhou2021patch2pix}, \cite{yu2023peal}, \cite{bai2021pointdsc}, \cite{gojcic2019perfect}), registration is achieved through iterative optimization between correspondence search and transformation estimation. However, these methods are highly sensitive to noise and density variations and are typically supervised. The second category, point-to-distribution, maps points to probability distributions and estimates transformations through distribution alignment or clustering \cite{mu2024colorpcr}, \cite{evangelidis2017joint}, \cite{huang2022unsupervised}. Although these methods are unsupervised, the iterative process can be time-consuming. Recent studies have adopted a coarse-to-fine approach \cite{zhang20233d}, \cite{huang2021predator}, achieving advanced performance. In this work, we follow this approach and focus on improving registration accuracy through a closer integration of color and geometric information (including color and distribution information).
\subsection{Point Cloud Feature Extraction}
Recently, due to the maturity of image matching methods, new paradigms for 3D point cloud feature extraction have emerged. PointNet performs deep learning-based feature extraction using graph convolution \cite{qi2017pointnet} and point convolution kernels \cite{qi2017pointnet++}. Notably, feature extraction using KPConv-FPN \cite{lin2017feature}, \cite{thomas2019kpconv} has become a mainstay. To enhance the use of color information, PEAL \cite{yu2023peal} extracts RGB color from images, but it faces the problem of color information loss. Based on this, ColorPCR \cite{mu2024colorpcr} integrates color information through the CEFE module and achieves better performance. However, these methods model color information separately. Inspired by the latest 3DGS technology in scene reconstruction, we consider tighter contextual relationships between color (opacity) and geometric information, enabling a Geometric-Color invariant representation. Therefore, we innovatively introduce 3DGS into point cloud registration and propose the Geometric-3DGS module. Through a color depth encoder, we extract deep color information and construct a global geographic relationship between point cloud color and geometric information to achieve fast registration in the coarse alignment stage.
\subsection{Registration With Color Features}
Some methods implicitly utilize color information to first detect keypoints or use 2D-3D multimodal learning \cite{el2021bootstrap}, \cite{wang2022improving}, \cite{yuan2023pointmbf}, \cite{zhang2022pcr}. PEAL detects overlaps in 2D images and then transfers them to a 3D registration network. Additionally, ColorPCR uses multi-stage color processing in the global registration process to utilize color information. The ICP algorithm \cite{park2017colored} and its variants (such as 4DICP) \cite{men2011color} increase the original point cloud feature dimensions to incorporate color information into geometric registration optimization. We follow this idea and use HSV and LAB to represent different color spaces \cite{liang2025control}, \cite{zhou2024blind}. Due to the differentiable rendering properties of 3DGS \cite{wu20244d}, \cite{yu2024mip}, \cite{luiten2024dynamic}, we adaptively fuse color information to achieve fast and stable fine registration of point clouds.
\section{Method}
\textbf{Problem statement.} Suppose we have two point clouds representing the target point cloud $P=\{p_i{\in}\mathbb{R}^3|i=1,...,N\}$ and the source point cloud $Q=\{q_i{\in}\mathbb{R}^e|i=1,...,M\}$, and their corresponding color information can be represented as $P_c=\{p_{c_i}\in[0,1]^3|i=1,...,N\}$ and $Q_c=\{q_{c_i}\in[0,1]^3|i=1,...,M\}$, respectively. The main objective of point cloud registration is to estimate a rigid transformation, represented by \( T = \{R, t\} \), where \( R \in SO(3) \) is the 3D rotation matrix and \( t \in \mathbb{R}^3 \) is the translation vector, such that the point clouds \( P \) and \( Q \) align through this transformation. The true correspondences between these point clouds are represented by the set \( C^* \), which is initially unknown. Therefore, we can optimize the following objective to solve for the rigid transformation:
\begin{equation}
	\min_{R, t} \sum_{(p_{x_i}^*, q_{y_i}^*) \in C^*} \| R \cdot p_{x_i}^* + t - q_{y_i}^* \|_2^2.
	\label{eq1}
\end{equation}

\begin{figure}[ht]
	\centering   
	\includegraphics[width=\linewidth]{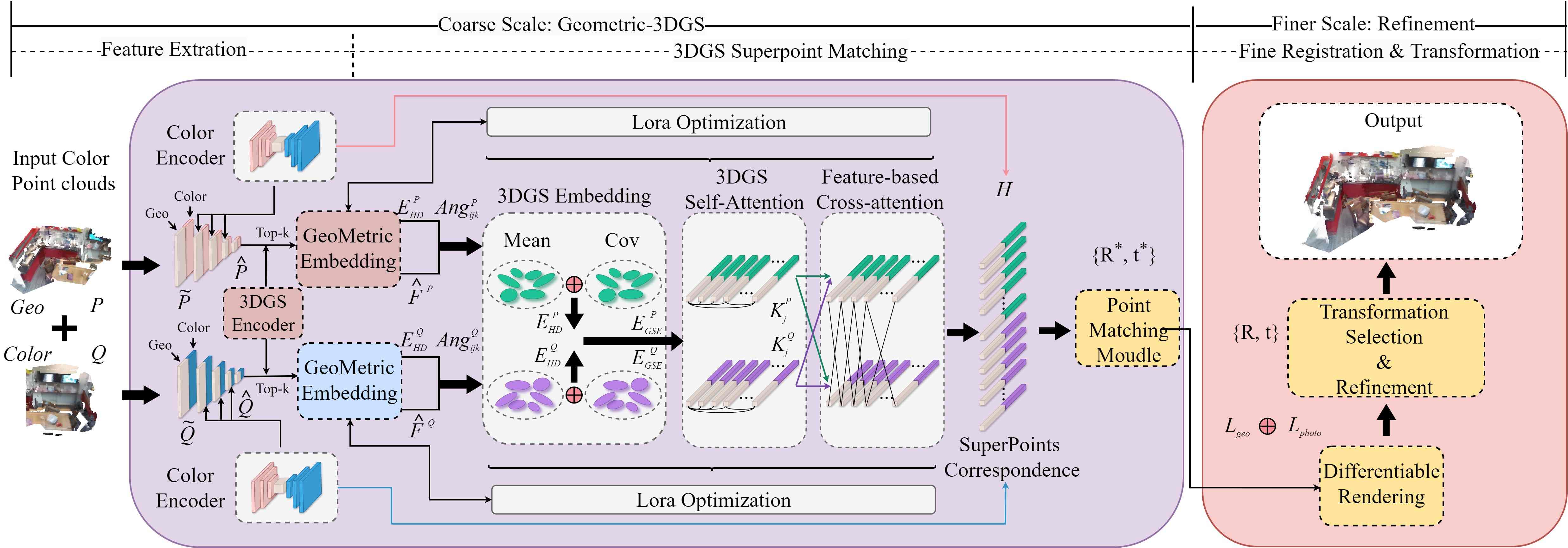}
	\caption{Pipeline. The entire network backbone is divided into coarse and fine scales. The feature extraction module extracts and integrates geometric and color information from the input point clouds \(P\) and \(Q\) using the color encoder and geometric encoder, producing superpoint representations \(\hat{P}\) and \(\hat{Q}\). The 3DGS Superpoint Matching Module identifies correspondences through 3DGS embeddings and self-attention mechanisms. LORA optimization is applied to reduce computational complexity and integrate both geometric and color information effectively. Finally, the Fine Registration \& Transformation Module refines the registration by performing differentiable rendering after coarse registration, optimizing the rigid transformation parameters \(\{R^*, t^*\}\).}
	\label{fig2}
\end{figure}
Our pipeline is shown in Fig.~\ref{fig2}. We represent the dense points and colors of the target point cloud as \( P \in \mathbb{R}^{|P|\times3} \) and \( P_c \in [0, 1]^{|P|\times3} \), respectively. The coarsest level of the dense points is represented as \( \tilde{P} \in \mathbb{R}^{|\tilde{P}|\times3} \) and \( \tilde{P_c} \in [0, 1]^{|\tilde{P}|\times3} \), and the finest level of the point cloud (superpoints) is represented as \( \hat{P} \in \mathbb{R}^{|\hat{P}|\times3} \) and \( \hat{P_c} \in [0, 1]^{|\hat{P}|\times3} \). Similarly, the source point cloud is represented in the same way.
\subsection{Coarse Registration With Color Features}
\subsubsection{Color Encoder Module}
We design a dedicated color encoder module to inject effective color information into point cloud features (see Fig. ~\ref{fig2}). The rationale for using this encoder is twofold: (1) typical geometry-color fusion is often performed by simple addition, which cannot effectively handle noise in color information; (2) direct color feature extraction across different levels of feature extraction leads to reliance on low-level features. The encoder first inputs the three-channel color vector $F_c \in \mathbb{R}^3$ into a multi-layer perceptron (MLP). This MLP processes the color information, fully encodes it, and extracts deep color features, which are then bounded in the [0, 1] range through a sigmoid function to produce normalized color features $F_C^{\prime}$. The noise-robust color mapping is as follows:

\begin{equation}
	F_C^{\prime}=\delta(LN(W_3\cdot{\delta(LN(W_2\cdot(\delta(LN(W_1\delta)))))})),
	\label{eq2}
\end{equation}

where $W_1$, $W_2$, and $W_3 \in \mathbb{R}^{d_{\text{in}} \times d_{\text{out}}}$ are learnable weights, $d_{\text{in}}$ and $d_{\text{out}}$ are input and output dimensions, $LN(\cdot)$ denotes layer normalization, and $\delta$ is the activation function. This process maps the original color to a decoupled high-dimensional space, effectively removing noise interference.

Based on this, we introduce a learned scalar weight $\alpha = \delta(\omega)$, where $\omega$ represents the parameter, to adaptively fuse the geometric and color features. During training, this weight dynamically adjusts the fusion process by balancing the contribution of geometric features $F_g$ and color information $F_c^{\prime}$. The final enhanced feature $F_{\text{enh}}$ is obtained as follows:
\begin{equation}
	F^{l}_{\text{enh}} = 
	\begin{cases} 
		F^{l}_g \oplus (\alpha \cdot F^{\text{color}}_c), & l < L \\ 
		F^{l}_g \oplus F^{\text{color}}_c, & l = L
	\end{cases} ,
	\label{eq3}
\end{equation}

where $F_g$ is the original geometric feature, $F_c^{\text{color}} \in \mathbb{R}^3$ is the color space vector (HSV, LAB) processed by the MLP, and $\oplus$ represents channel concatenation. We use this color encoder in feature extraction at different levels. At the final level, we only use feature concatenation without the color encoder. This design injects color information while preserving the original geometry and other feature information, ensuring sufficient multimodal representation for downstream tasks.
\subsubsection{Geometric-3DGS Module}
The Geometric-3DGS module mainly consists of three components: the 3DGS encoder, attention with 3DGS embeddings, and Gaussian superpoint registration, as shown in Fig.~\ref{fig2}. To improve the robustness and accuracy of the coarse registration stage, this module deeply integrates geometric and color features. Directly parameterizing all point cloud information with 3DGS at the beginning would result in a large amount of redundant parameters and very high memory requirements. Therefore, the core of this module is to 3DGS encode the local neighborhood information during point cloud downsampling, and use 3DGS embeddings and attention mechanisms to focus on relevant features in the 3DGS encoding. This module implements feature extraction at different granularities and provides a global, transformation-invariant geometric-color representation, which is crucial for fine point cloud registration. \\
\textbf{3DGS encoder.} We propose a Dual-Modal Color Encoder (DMCE), which is responsible for transforming local neighborhood patches in the point cloud into more robust and rich representations, thus capturing both geometric and color features. This method uses Gaussian distribution-based representations to model the local neighborhood structure and the relationships between points. The specific steps of the 3DGS encoder are as follows:
We calculate the covariance matrix of each local neighborhood in the point cloud, which can capture the local geometric structure. This matrix considers the relative distances between points within the neighborhood to enhance the representation of local surface directions. Specifically, for a key point $P_i$ with a neighborhood $N_i$, we first compute the differentiable covariance matrix:
\begin{equation}
	{\sum}_i = Ro(r_i) \cdot diag(\exp(s_i)) \cdot Ro(r_i)^T + \lambda n_in_i^T,
	\label{eq4}
\end{equation}
where $Ro(\cdot) \in SO(3)$ represents the rotation matrix, $r_i \in \mathbb{R}^3$ is the rotation quaternion, $s_i \in \mathbb{R}^3$ is the logarithmic scale vector, $n_i$ is the estimated normal vector, and $\lambda$ controls the strength of the normal vector, with $\lambda \in [0.01, 0.1]$. This covariance matrix captures the geometric features of the local region of the point cloud. Finally, we perform top-k ($k\in[2, 5]$) fusion between the color features from the color encoder and the 3DGS parameters. By combining position, covariance matrix, color, and transparency, we construct the 3DGS feature vector:
\begin{equation}
	F_{3DGS}^i = Top_{k_i}(Concat[{\mu}_i, vec({\Sigma}_i), {\alpha}_i, F_{enh}^i])
	\label{eq5}
\end{equation}
where $\mu _ i$ is the Gaussian position center, and $\alpha _ i$ represents transparency.\\
\textbf{Attention with 3DGS embeddings.} In this part, we implement 3DGS position embeddings to obtain globally invariant geometric-color fusion encodings. Finally, based on the GeoTransformer, we use self-attention and cross-attention to focus on the color information in the point cloud structure and guide superpoint registration. In both indoor and outdoor scenes, uncertainties such as sensors, lighting, and natural weather conditions often introduce noise or missing data in color sampling. To mitigate the impact of these color noises on global features, we aim to differentiate the parts that are heavily affected by noise. For indoor scenes, we use the HSV color space, and for outdoor scenes, we use the LAB color space. This ensures the most stable color information embedding, thus enabling robust structural position encoding. The specific process for the 3DGS position embedding is in Appendix \ref{appendixA2}\\
\textbf{Superpoint registration with 3DGS.} To quickly register the point cloud information from Geometric-3DGS during the coarse registration stage, we use the ICP algorithm to align the Gaussian distributions. By minimizing the 3DGS distance between the source and target point clouds $p_i, p_j$,  we iteratively optimize the rigid transformation between $p_i, p_j$ using least squares until convergence. We define the generalized distance metric as $d_{ij} = \left\| \mu_i^s - \mu_j^t \right\|_2^2 + \lambda_d \left\| \Sigma_i^s - \Sigma_j^t \right\|_F$, where the parameter $\lambda_d$ is adjustable within the range of [0.05, 0.2], $\mu_i^s$, $\mu_j^t$ are the means of the Gaussians in the source and target point clouds, and $\Sigma_i^s$, $\Sigma_j^t$ are the covariances. Based on this definition, we give the associated probability matrix $A \in R^{N \times M}$, which measures the confidence of the match between $p_i$, $p_j$:
\begin{equation}
	A_{ij} = \frac{\exp(-\gamma d_{ij})}{\sum_{k=1}^{M} \exp(-\gamma d_{ik})},
	\label{eq6}
\end{equation}
where $\gamma$ is the sensitivity control parameter, adjustable within the range of [1, 3], and this probability matrix is normalized using softmax. By combining the generalized distance and the probability matrix, we construct the objective to be optimized based on Mahalanobis distance:
\begin{equation}
	\min_{R \in \text{SO}(3), t \in \mathbb{R}^3} \sum_{i=1}^{N} \sum_{j=1}^{M} A_{ij} \left\| R (\mu_i^s - \bar{\mu}^s) + t - (\mu_j^t - \bar{\mu}^t) \right\|_{\Sigma_j^{-1}}^{2},
	\label{eq7}
\end{equation}
where $\mu_i^s$, $\mu_j^t$ represent the weighted centroids of the source and target point clouds, respectively, $ \bar{\mu}^s = \frac{1}{N} \sum_{i} A_{ij} \mu_i^s$, $\quad \bar{\mu}^t = \frac{1}{N} \sum_{i} A_{ij} \mu_i^t$. $\left\| \cdot \right\|$ denotes Mahalanobis distance, with ${\Sigma}_j^{-1}$ used to down-weight regions where the covariance computation has large errors. This objective can be solved using SVD decomposition, resulting in:
\begin{equation}
	H = \sum_{ij} A_{ij} \Sigma_j^{-1} (\mu_i^s - \bar{\mu}^s)(\mu_j^t - \bar{\mu}^t)^T,
	\label{eq8}
\end{equation}
\begin{equation}
	U, S, V = {SVD}(H), \quad R^* = VU^T, \quad t^* = \bar{\mu}^t - R^* \bar{\mu}^s.
	\label{eq9}
\end{equation}
Once the transformation values during training are smaller than a minimum tolerance, the 3DGS registration is achieved.\\
\textbf{LORA optimization.} In the Geometric-3DGS module, the 3DGS parameterization introduces a large amount of high-dimensional data, which may result in significant computational and storage burdens, especially in large-scale point cloud scenarios. Therefore, ` By using the LORA module, we convert the high-dimensional 3DGS embeddings into a low-rank form, allowing the model to remain efficient and accurate without sacrificing its capacity.
\subsection{Fine Registration With Photometric Optimization}
To improve point cloud registration accuracy, we propose a fine registration method based on photometric optimization. After coarse registration, we optimize point cloud alignment by rendering the 3DGS of the target and source point clouds under the new pose and minimizing the weighted photometric loss. Traditional photometric optimization methods typically only consider the L1 distance, which fails to effectively handle noise and lighting variations. To address this, we introduce a weighted photometric loss, assigning different weights to each pixel to enhance robustness and reduce the effects of uneven lighting and noise. Our photometric loss is calculated as follows:
\begin{equation}
	L_{photo} = \sum_i \omega_i \left\| \hat{f}(G_1, \bar{C}) - \hat{f}(G_2, \bar{C}) \right\|^2, \quad \omega_i = \exp(-\gamma_d \cdot d_i),
	\label{eq10}
\end{equation}
where $\hat{f}(G_1, \bar{C})$, $\hat{f}(G_2, \bar{C})$ represent the rendered results of the target and source point clouds under the new pose $\bar{C}$, $G_1$ and $G_2$represent the 3DGS from the coarse registration, $\omega_i$ represents the pixel weight factor, and $d_i$ is the pixel distance between the source and target point clouds. The sensitivity control parameter $\gamma_d$ adjusts the influence of distance on registration, adjustable within the range of [1, 3]. Using differentiable rendering, we backpropagate the loss to the transformation parameters $R^*$, $t^*$ and update them with gradient descent. This approach enhances accuracy, particularly in key regions, by combining geometric loss with the weighted photometric loss for precise point cloud alignment, as shown:
\begin{equation}
	L_{total} = L_{geo} + \lambda_p L_{photo}.
	\label{eq11}
\end{equation}
where $L_{geo}$ is the geometric registration loss, and $\lambda_p \in [0.1, 1]$ is a weight hyperparameter that adjusts the balance between geometric and photometric losses. We provide the detailed proof of the convergence of the joint photometric optimization in the Appendix \ref{appendixA1}.
\section{Experiments}
To validate the performance of the GeGS-PCR model, we evaluate it on the indoor benchmarks Color3DMatch (C3DM) and Color3DLoMatch (C3DLM), as well as our colorized outdoor ColorKitti (The specific data construction process, including the detailed steps and settings for preparing the datasets, can be found in Appendix \ref{appendixA4} and \ref{appendixA5}. ) odometry benchmark. Each point cloud in these datasets includes an RGB color value.
\begin{table}[hb]
	\caption{Evaluation results on C3DM and C3DLM. \#Samples in the table represents the number of correspondences selected by RANSAC.}
	\label{tab1}
	\centering
	\resizebox{\textwidth}{!}{  
		\begin{tabular}{l|ccccc|ccccccc}
			\toprule
			& \multicolumn{5}{c|}{C3DM} & \multicolumn{5}{c}{ C3DLM} \\
			\#Samples & 5000 & 2500 & 1000 & 500 & 250 & 5000 & 2500 & 1000 & 500 & 250 \\
			\midrule
			\multicolumn{11}{c}{Feature Matching Recall (\%) $\uparrow$} \\
			\hline
			CoFiNet \cite{yu2021cofinet} & 98.1 & 98.3 & 98.1 & 98.2 & 98.3 & 83.1 & 83.5 & 83.3 & 83.1 & 82.6 \\
			GeoTransformer \cite{qin2022geometric} & 97.9 & 97.9 & 97.9 & 97.9 & 97.6 & 88.3 & 88.6 & 88.8 & 88.6 & 88.3 \\
			PEAL \cite{yu2023peal} & 99.0 & 99.0 & 99.1 & 99.1 & 98.8 & 91.7 & 92.4 & 92.5 & 92.9 & 92.7 \\
			ColorPCR \cite{mu2024colorpcr} & 99.5 & 99.5 & 99.5 & 99.5 & 99.5 & 96.5 & 96.5 & 97.0 & 97.0 & 96.7 \\
			GeGS-PCR (ours) & \textbf{99.5} & \textbf{99.6} & \textbf{99.5} & \textbf{99.7} & \textbf{99.6} & \textbf{97.6} & \textbf{97.4} & \textbf{97.1} & \textbf{97.2} & \textbf{97.0} \\
			\midrule
			\multicolumn{11}{c}{Inlier Ratio (\%) $\uparrow$} \\
			\hline
			CoFiNet \cite{yu2021cofinet} & 49.8 & 51.2 & 51.9 & 52.2 & 52.2 & 24.4 & 25.9 & 26.7 & 26.8 & 26.9 \\
			GeoTransformer \cite{qin2022geometric} & 71.9 & 75.2 & 76.0 & 82.2 & 85.1 & 43.5 & 45.3 & 46.2 & 52.9 & 57.7 \\
			PEAL \cite{yu2023peal} & 72.4 & 79.1 & 84.1 & 86.1 & 87.3 & 45.0 & 50.9 & 57.4 & 60.3 & 62.2 \\
			ColorPCR \cite{mu2024colorpcr} & 75.0 & 80.5 & 84.7 & 86.5 & 87.8 & 51.2 & 56.6 & 63.1 & 66.0 & 68.0 \\
			GeGS-PCR (ours) & \textbf{76.3} & \textbf{82.4} & \textbf{86.3} & \textbf{86.6} & \textbf{89.1} & \textbf{53.4} & \textbf{58.7} & \textbf{66.9} & \textbf{69.7} & \textbf{70.3} \\
			\midrule
			\multicolumn{11}{c}{Registration Recall (\%) $\uparrow$} \\
			\hline
			CoFiNet \cite{yu2021cofinet} & 89.3 & 88.9 & 88.4 & 87.4 & 87.0 & 67.5 & 66.2 & 64.2 & 63.1 & 61.0 \\
			GeoTransformer \cite{qin2022geometric} & 92.0 & 91.8 & 91.8 & 91.4 & 91.2 & 75.0 & 74.8 & 74.2 & 74.1 & 73.5 \\
			PEAL \cite{yu2023peal} & 94.6 & 93.7 & 93.7 & 93.9 & 93.4 & 81.7 & 81.2 & 80.8 & 80.4 & 80.1 \\
			GeoTransformer+MAC \cite{qin2022geometric} & 95.7 & 95.7 & 95.2 & 95.3 & 94.6 & 78.9 & 78.7 & 78.2 & 77.7 & 76.6 \\
			ColorPCR \cite{mu2024colorpcr} & 97.5 & 96.5 & 97.0 & 96.4 & 96.5 & 88.9 & 88.5 & 88.1 & 86.5 & 85.0 \\
			GeGS-PCR (ours) & \textbf{97.9} & \textbf{97.6} & \textbf{97.5} & \textbf{96.7} & \textbf{97.6} & \textbf{90.7} & \textbf{90.2} & \textbf{90.4} & \textbf{90.0} & \textbf{89.8} \\
			\hline
		\end{tabular}
	}
\end{table}
\subsection{Indoor Benchmarks: Color3DMatch \& Color 3DLoMatch}
\begin{table}[h]
	\caption{Registration results w/o RANSAC on C3DM and C3DLM.  The model is the time for feature extraction, while the pose time is for transformation estimation}
	\label{tab2}
	\centering
	\resizebox{\textwidth}{!}{
		\begin{tabular}{l |c| c| c c| c c c}
			\toprule
			\multirow{2}{*}{Model} & \multirow{2}{*}{Estimator} & \multirow{2}{*}{\#Sample} & \multicolumn{2}{c|}{RR (\%)$\uparrow$} & \multicolumn{3}{c}{Time (s)$\downarrow$} \\
			& & & C3DM & C3DLM & Model & Pose & Total \\
			\midrule
			FCGF\cite{choy2019fully} & RANSAC-50k & 5000 & 85.1 & 40.1 & 0.052 & 3.326 & 3.378 \\
			D3Feat\cite{bai2020d3feat} & RANSAC-50k & 5000 & 81.6 & 37.2 & \textbf{0.024} & 3.088 & 3.112 \\
			SpinNet\cite{ao2021spinnet} & RANSAC-50k & 5000 & 88.6 & 59.8 & 60.248 & \textbf{0.388} & 60.636 \\
			Predator\cite{huang2021predator} & RANSAC-50k & 5000 & 89.0 & 59.8 & \underline{0.032} & 5.120 & 5.152 \\
			CoFiNet\cite{yu2021cofinet} & RANSAC-50k & 5000 & 89.3 & 67.5 & 0.115 & 1.807 & 1.922 \\
			GeoTransformer\cite{qin2022geometric} & RANSAC-50k & 5000 & 92.0 & 75.0 & 0.075 & \underline{1.558} & \textbf{1.633} \\
			PEAL\cite{yu2023peal} & RANSAC-50k & 5000 & 94.6 & 81.7 & 0.089 & 1.776 & 1.865\\
			ColorPCR\cite{mu2024colorpcr} & RANSAC-50k & 5000 & 97.5 & 88.2 & 0.083 & 1.629 & 1.712\\
			GeGS-PCR(ours) & RANSAC-50k & 5000 & \textbf{97.9} & \textbf{90.7} & 0.082 & 1.618 & \underline{1.700} \\
			\midrule
			CoFiNet\cite{yu2021cofinet} & RANSAC-free & all & 87.6 & 64.8 & 0.115 & \underline{0.028} & 0.143 \\
			GeoTransformer\cite{qin2022geometric} & RANSAC-free & all & 91.5 & 74.0 & \underline{0.075} & \textbf{0.013} & \textbf{0.088} \\
			PEAL\cite{yu2023peal} & RANSAC-free & all & 94.3 & 78.8 & 0.089 & 0.034 & 0.133\\
			ColorPCR\cite{mu2024colorpcr} & RANSAC-free & all & 97.3 & 88.3 & 0.083 & 0.046 & 0.129\\
			GeGS-PCR & RANSAC-free & all & \textbf{97.5} & \textbf{89.1} & \textbf{0.082} & 0.032 & \underline{0.124}\\
			\bottomrule
		\end{tabular}
	}
\end{table}
\textbf{Correspondence results.} We compared GeGS-PCR with several SOTA methods (metrics in Appendix~\ref{appendixA3}). Most methods use RANSAC, and we followed the same approach for consistency. Key results are shown in Table~\ref{tab1} (additional results in the Appendix~\ref{appendixA3}). GeGS-PCR outperforms ColorPCR with 99.5\% FMR on C3DM and 97.6\% on C3DLM. For IR, it reaches 89.1\% on C3DM and 70.3\% on C3DLM, surpassing ColorPCR by 1.3\% on C3DM and 2.3\% on C3DLM. In RR, GeGS-PCR achieves 97.9\% on C3DM and 90.7\% on C3DLM, outperforming ColorPCR by 0.4\% on C3DM and 4.2\% on C3DLM. These results demonstrate significant improvement over baseline methods, with GeGS-PCR excelling, especially in low-overlap scenarios.
\begin{table}[h]
	\caption{Performance of ablation experiments}
	\label{tab3}
	\centering
	\resizebox{\textwidth}{!}{
		\begin{tabular}{l|cccc|cccc}
			\toprule
			& \multicolumn{4}{c|}{C3DM} & \multicolumn{4}{c}{C3DLM}\\ 
			Overlap & PIR(\%) & FMR(\%) & IR(\%) & RR(\%) & PIR(\%) & FMR(\%) & IR(\%) & RR(\%) \\
			\midrule
			(a) w/o differentiable rendering & 71.7 & 98.0 & 59.1 & 90.7 & 45.6 & 85.8 & 40.1 & 72.1 \\
			(b) w/o 3DGS & 75.9 & 97.9 & 65.6 & 91.0 & 50.6 & 87.1 & 42.5 & 73.1 \\
			(c) w/o Geometric-3DGS & 80.1 & 97.9 & 70.4 & 91.3 & 51.5 & 88.2 & 42.7 & 73.5 \\
			(d) w/o color encoder & 82.1 & 97.7 & 70.3 & 91.5 & 53.9 & 88.1 & 43.3 & 74.0 \\
			(e) w/o color & 83.8 & 98.0 & 70.7 & 92.7 & 54.1 & 88.4 & 43.5 & 74.2 \\
			(f) w/o LoRA & \textbf{92.0} & 99.3 & 86.3 & 97.5 & \textbf{63.8} & 97.4 & 59.9 & 90.5 \\
			(g) Geometric-3DGS(Full) & 92.0 & 99.6 & 82.4 & 97.6 & 63.9 & 97.4 & 58.7 & 90.2 \\
			\bottomrule
		\end{tabular}
	}  
\end{table}
\begin{figure}[h]
	\centering
	\includegraphics[width=\linewidth]{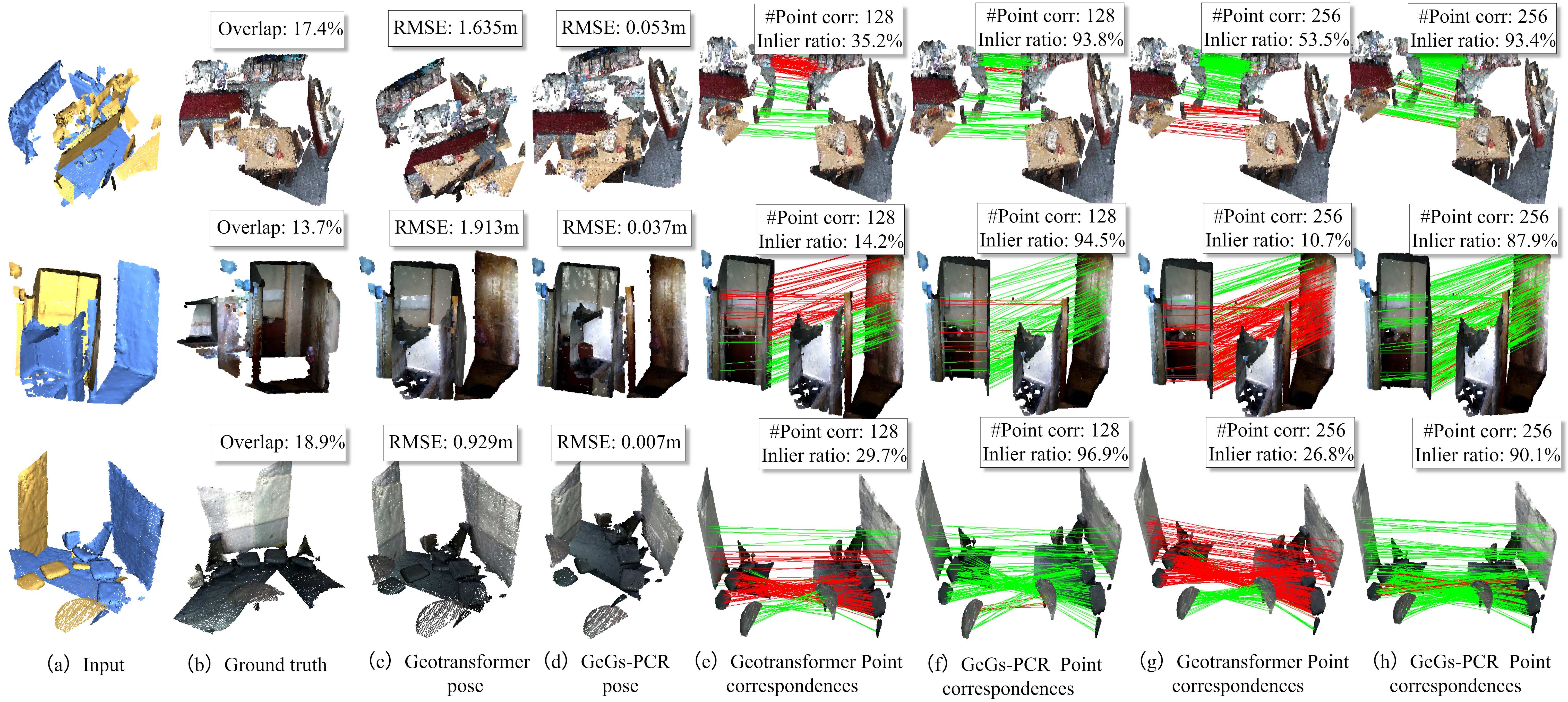} 
	\caption{Registration performance with GeGS-PCR and Geometric self-attention.}
	\label{fig3}
\end{figure}
\begin{table}[h]
	\caption{Ablation results based on ColorPCR baseline}
	\label{tab4}
	\centering
	\resizebox{\textwidth}{!}{
		\begin{tabular}{l|cccc|cccc}
			\toprule
			& \multicolumn{4}{c|}{C3DM} & \multicolumn{4}{c}{C3DLM}\\ 
			Method & PIR(\%) & FMR(\%) & IR(\%) & RR(\%) & PIR(\%) & FMR(\%) & IR(\%) & RR(\%) \\
			\midrule
			(a) ColorPCR (baseline) & 89.2 & 99.5 & 80.5 & 96.5 & 62.7 & 96.5 & 56.6 & 88.5 \\
			(b) w/o ColorEncoder & 89.4 & 99.5 & 80.6 & 96.6 & 62.8 & 96.6 & 56.8 & 88.6 \\
			(c) w/o 3DGS & 89.5 & 99.6 & 80.7 & 96.7 & 63.0 & 96.7 & 56.9 & 88.8 \\
			(d) w/o differentiable rendering & 89.6 & 99.5 & 80.8 & 96.8 & 63.1 & 96.8 & 57.0 & 88.9 \\
			(e) w/o color & 86.1 & 97.9 & 77.3 & 92.7 & 55.2 & 89.8 & 46.3 & 77.9 \\
			(f) w/o Geometric PE & 88.8 & 99.3 & 80.2 & 96.3 & 62.4 & 96.0 & 56.4 & 88.0 \\
			(g) w/o LoRA & 88.9 & 99.3 & 80.1 & 96.2 & 62.4 & 96.1 & 56.3 & 88.4 \\
			(h) Geometric-3DGS (all) & \textbf{90.0} & \textbf{99.6} & \textbf{82.4} & \textbf{97.6} & \textbf{63.9} & \textbf{97.4} & \textbf{58.7} & \textbf{90.2} \\
			\bottomrule
		\end{tabular}
	}  
\end{table}\\
\textbf{Registration results.} As shown in Table~\ref{tab2}, GeGS-PCR outperforms both RANSAC and RANSAC-free methods. For RANSAC-based methods, GeGS-PCR achieves 97.9\% RR on C3DM and 90.7\% on C3DLM, surpassing ColorPCR, with a total processing time of 1.703s, second only to D3Feat (1.712s). GeGS-PCR also achieves the best pose estimation time of 0.072s. For RANSAC-free methods, GeGS-PCR reaches 96.9\% RR on C3DM and 89.1\% on C3DLM, outperforming ColorPCR, with a total time of 0.124s, second only to GeoTransformer (0.088s). However, GeGS-PCR surpasses GeoTransformer in both RR and pose time, demonstrating its superior precision and speed.\\
\begin{wraptable}{r}{0.65\textwidth}
	\caption{Registration results w/o RANSAC on Kitti}
	\label{tab5}
	\centering
	\resizebox{0.65\textwidth}{!}{
		\begin{tabular}{l|c|c|c}
			\toprule
			Model & RTE (cm) & RRE (°) & RR (\%) \\
			\midrule
			3DFeat-Net \cite{yew20183dfeat} & 25.9 & 0.25 & 96.0 \\
			FCGF \cite{choy2019fully} & 9.5 & 0.30 & 96.6 \\
			D3Feat \cite{bai2020d3feat} & 7.2 & 0.30 & 99.8 \\
			SpinNet \cite{ao2021spinnet} & 9.9 & 0.47 & 99.1 \\
			Predator \cite{huang2021predator} & 6.8 & 0.27 & 99.8 \\
			CoFiNet \cite{yu2021cofinet} & 8.2 & 0.41 & 99.8 \\
			GeoTransformer \cite{qin2022geometric} & 7.4 & 0.27 & 99.8 \\
			GeGS-PCR (ours, RANSAC-50k) & \textbf{6.3} & \textbf{0.16} & \textbf{99.9} \\ 
			\midrule
			FMR \cite{huang2020feature} & $\sim$66 & 1.49 & 90.6 \\
			DGR \cite{choy2020deep} & $\sim$32 & 0.37 & 98.7 \\
			HRegNet \cite{lu2021hregnet} & $\sim$12 & 0.29 & 99.7 \\
			GeoTransformer (LGR) & 6.8 & 0.24 & 99.8 \\
			GeGS-PCR (ours, LGR) & \textbf{5.7} & \textbf{0.13} & \textbf{99.9} \\ 
			\bottomrule
		\end{tabular}
	}
\end{wraptable}
\textbf{Ablation experiments.} Table~\ref{tab3} presents the ablation experiment results, analyzing the impact of each module on registration performance. Removing the color component (row e) reduces performance, particularly on C3DLM. Without the color encoder (row d), performance drops slightly, especially in FMR. Excluding the Geometric-3DGS module (row c) decreases PIR and IR, and removing 3DGS (row b) further lowers registration recall (RR). The removal of differentiable rendering (row a) significantly affects IR and PIR, with decreases of 7.1\% and 9.5\% in PIR and 6.5\% and 8.1\% in IR for C3DM and C3DLM, respectively. This highlights the importance of differentiable rendering for improving precision and inlier ratio. It helps the model utilize color information, improving point cloud alignment and boosting IR and PIR. In addition, removing LoRA optimization (row f) leads to a slight drop in registration performance, particularly in IR and RR, indicating that LoRA mainly accelerates convergence and provides a modest yet consistent improvement in accuracy while preserving efficiency. More detailed ablation analysis is shown in Appendix~\ref{appendixA5}. \\
\textbf{Baselines ablation experiments} Table~\ref{tab4} reports the ablation results based on the ColorPCR baseline. Overall, each module contributes to performance improvements to varying degrees. Removing color information (row e) causes the most significant degradation, with PIR, IR, and RR dropping notably on both C3DM and C3DLM, highlighting the critical role of color features in low-overlap scenarios. In contrast, removing the color encoder (row b) or geometric positional encoding (row f) only leads to minor fluctuations, suggesting these modules play supportive roles. Excluding 3DGS (row c) or differentiable rendering (row d) results in moderate decreases, particularly in IR and RR, indicating their importance for fine-grained alignment and convergence. Removing LoRA (row g) slightly reduces performance, confirming its role in accelerating convergence and providing consistent accuracy gains while maintaining efficiency. Finally, the full Geometric-3DGS model (row h) achieves the best results across all metrics, with PIR reaching 90.0\% and IR improving to 82.4\% on C3DM, and PIR reaching 63.9\% and IR improving to 58.7\% on C3DLM, demonstrating the effectiveness of the proposed modules in synergy. To clarify metric repetition, we ran independent ablations on ColorPCR (Table~\ref{tab4}) controlling module order and data synergy. Results show consistent gains even on weaker baselines, proving our module design is sound, generalizable, and optimizable.
\textbf{Qualitative Results.} Fig.~\ref{fig3} shows the comparison of registration results using geometric attention and our 3DGS self-attention in scenes with low overlap and geometric feature overlap. Geometric attention produces a large dispersive effect, while 3DGS self-attention is able to find consistent correspondences for low-overlap features, significantly improving the inlier ratio and resulting in more accurate registration. The visualization results show that 3DGS self-attention can accurately identify correspondences in low-overlap areas and reject incorrect matches.
\begin{figure}[h]
	\centering
	\includegraphics[width=\linewidth]{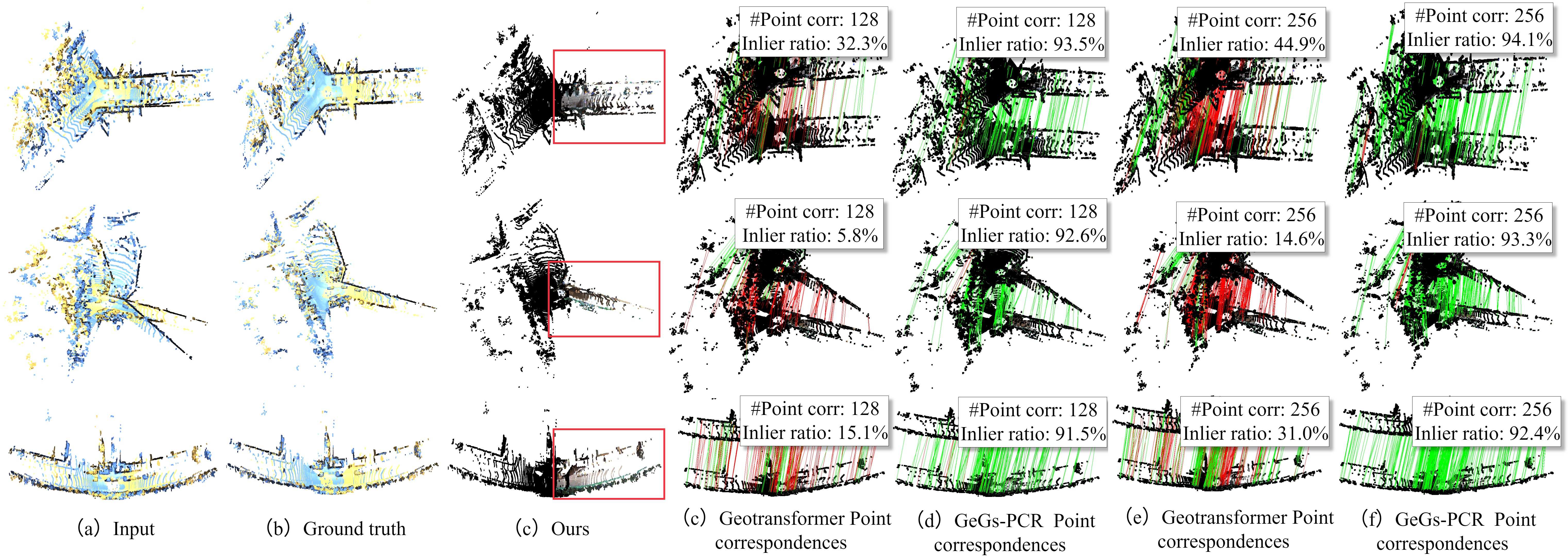}
	\caption{Registration performance with GeGS-PCR and Geometric self-attention.}
	\label{fig4}
\end{figure}
\subsection{Outdoor Benchmarks: ColorKitti}
\textbf{Registration results.} Based on previous research settings \cite{ao2021spinnet, qin2022geometric, yu2021cofinet}, we introduced the Registration Recall (${RR_{Outdoor}}$) metric (defined in Appendix~\ref{appendixA3}). In Table~\ref{tab5}, we first compare our GeGS-PCR model with RANSAC-based methods. Additionally, we compare it with methods that do not use RANSAC. The results show that GeGS-PCR performs comparably, with significant improvements in some metrics. Using RANSAC, GeGS-PCR achieves 99.9\% Registration Recall (RR), demonstrating excellent performance. Even without RANSAC, GeGS-PCR maintains a 99.9\% RR, and achieves the best RTE and RRE values.\\
\textbf{Qualitative results. } Fig.~\ref{fig4} shows the comparison of registration results using geometric attention and our 3DGS self-attention in low-overlap scenes. The red boxes in the ground truth highlight the colored regions. The results demonstrate that GeGS-PCR achieves good registration in these regions under both high and low point correspondence conditions. Notably, GeGS-PCR produces almost no incorrect correspondences in the colored areas. While geometric attention causes significant dispersive effects, 3DGS self-attention consistently finds correct correspondences in low-overlap features, significantly improving the inlier ratio, especially in such regions. Further limitations and a comprehensive performance analysis can be found in Appendix~\ref{appendixA5} and Appendix~\ref{appendixA6}.
\section{Conclusion}
In this paper, we present GeGS-PCR, an innovative two-stage point cloud registration method that enhances registration accuracy and robustness by integrating geometric, color, and Gaussian distribution information. In the coarse registration phase, GeGS-PCR effectively extracts reliable features in low-overlap scenarios with incomplete geometric features by introducing the Geometric-3DGS module and a color encoder. Additionally, LoRA optimization is applied to reduce the complexity introduced by feature encoding. In the fine registration phase, differentiable rendering combined with photometric optimization loss further improves performance. The advantages of GeGS-PCR lie in the collaborative optimization of global and local structures. Through local Gaussian feature extraction, GeGS-PCR effectively suppresses noise interference and robustly fuses geometric and color features. GeGS-PCR is not only suitable for high-overlap point cloud registration but also offers an efficient solution for low-overlap tasks, such as autonomous driving and large-scale scene reconstruction.

\section{Acknowledgments}
This work was supported in part by National Natural Science Foundation of China No. 62088102 and No.62302381. The Authors are with the National Key Laboratory of Human-Machine Hybrid Augmented Intelligence, National Engineering Research Center of Visual Information and Applications, and Institute of Artificial Intelligence and Robotics, Xi'an Jiaotong University, Xi'an, Shaanxi, China.

\newpage
\small
\bibliographystyle{unsrt}
\bibliography{Manuscript}  


\newpage
\appendix

\section{Technical Appendices and Supplementary Material}
\subsection{Proof of photometric optimization}
\label{appendixA1}
To rigorously analyze the convergence of our joint photometric-geometric optimization, we formulate the camera pose optimization on the Special Euclidean group \( \text{SE}(3) \) manifold. Let the camera pose be parameterized as \( T \in \text{SE}(3) \), which can be expressed via the exponential map from its Lie algebra \( \xi \in \mathfrak{se}(3) \):
\begin{equation}
	T = \exp(\xi^\wedge), \quad \xi = [\omega, v]^T \in \mathbb{R}^6,
	\label{eq12}
\end{equation}
where \( \omega \in \mathbb{R}^3 \) and \( v \in \mathbb{R}^3 \) represent the angular and linear velocities, respectively. They can be derived from the rotation matrix $R$ and the translation matrix $T$. Under the assumptions of local convexity of the photometric loss \( L_{\text{photo}} \) near the optimal pose \( T^* \) and bounded gradient noise, the proposed joint loss \( L_{\text{total}} \) converges to a local minimum when optimized via Riemannian gradient descent on SE(3). The gradient of \( L_{\text{total}} \) on SE(3) is computed through the retraction map \( R_T \):
\begin{equation}
	\text{grad} \, L_{total} = R_T^{-1} \left( \frac{\partial L_{geo}}{\partial \xi} + \lambda \frac{\partial L_{photo}}{\partial \xi} \right).
	\label{eq13}
\end{equation}
where \( \xi \in \mathfrak{se}(3) \) is the Lie algebra of the transformation and \( R_T^{-1} \) is the retraction map that ensures the gradient is mapped correctly from the Lie algebra back to SE(3). Using the chain rule, the photometric gradient with respect to \( \xi \) is derived as:
\begin{equation}
	\frac{\partial L_{photo}}{\partial \xi} = \sum_i \omega_i \cdot \frac{\partial \| f(G_1) - f(G_2) \|^2}{\partial \Delta I} \cdot \frac{\partial \Delta I}{\partial T} \cdot \frac{\partial T}{\partial \xi},
	\label{eq14}
\end{equation}
where \( f(G_1) \) and \( f(G_2) \) represent the rendered results of the target and source point clouds under the new pose \( C \), \( \omega_i \) is the weight factor for each pixel, and \( \Delta I \) is the difference between the rendered and target images. The image gradient is:
\begin{equation}
	\frac{\partial \| f(G_1) - f(G_2) \|^2}{\partial \Delta I} = 2 \cdot (f(G_1) - f(G_2)),
	\label{eq15}
\end{equation}
We also calculate the rendering Jacobian and Lie Algebra Jacobian as:
\begin{equation}
	\frac{\partial \Delta I}{\partial T}, \quad \frac{\partial T}{\partial \xi},
	\label{eq16}
\end{equation}
The Lie algebra Jacobian follows the left perturbation model:
\begin{equation}
	\frac{\partial T}{\partial \xi} = \lim_{\delta \xi \to 0} \exp((\delta \xi)^\wedge) T - T_\delta,
	\label{eq17}
\end{equation}
Using the Łojasiewicz inequality on matrix Lie groups, we show that with learning rate \( \eta < \frac{1}{L} \) (where \( L \) is the Lipschitz constant), the gradient norm \( \| \text{grad} L_{\text{total}}^{(k)} \|_2^2 \) decreases monotonically until reaching a stationary point:
\begin{equation}
	\| \text{grad} \, L_{total}^{(k)} \|_2^2 \leq C (L_{total}^{(k)} - L_{total}^{(k+1)}).
	\label{eq18}
\end{equation}
This guarantees that the loss function will decrease monotonically and converge to a local minimum.
\subsection{Structure Embedding}
\label{appendixA2}
\textbf{3DGS embedding.} In this part, we implement 3DGS position embeddings to obtain globally invariant geometric-color fusion encodings. Finally, based on the GeoTransformer, we use self-attention and cross-attention to focus on the color information in the point cloud structure and guide superpoint registration. In both indoor and outdoor scenes, uncertainties such as sensors, lighting, and natural weather conditions often introduce noise or missing data in color sampling. To mitigate the impact of these color noises on global features, we aim to differentiate the parts that are heavily affected by noise. For indoor scenes, we use the HSV color space, and for outdoor scenes, we use the LAB color space. This ensures the most stable color information embedding, thus enabling robust structural position encoding. The specific process for the 3DGS position embedding is as follows:\\
\textbf{Gaussian embedding.} We add Gaussian embedding to capture geometric distribution information that is closely related to color information. This is particularly useful for invariance between point clouds, where the distribution of local regions is represented using the mean $\mu$ and covariance $\Sigma$ of 3DGS. The difference in the mean is computed as $\delta_{\mu_{ij}} = \left\| \mu_i - \mu_j \right\|_2$, where $\mu_i$, $\mu_j$ represent the local means in the source and target point clouds, respectively. The Frobenius norm difference in covariance is $\delta_{\Sigma_{ij}} = \left\| \Sigma_i - \Sigma_j \right\|_F$, where $\Sigma_i$, $\Sigma_j$ are the covariance matrices of the local neighborhoods in the two point clouds. The final Gaussian embedding is computed as:
\begin{equation}
	E_{GS} = PE \left( \frac{(\delta _{\Sigma_{ij}} + \delta _{\Sigma_{ij}}) \cdot \delta_{\mu_{ij}}}{\sigma_{GS}} \right) W_{GS},
	\label{eq19}
\end{equation}
where $\sigma_{GS}$ is the sensitivity hyperparameter for the Gaussian embedding, adjustable within the range of [0.01, 0.1], PE is the sine position embedding function, and $W_{GS} \in \mathbb{R}^{d_t \times d_t}$ is the matrix used for projection, with $d_t$ representing the output dimension. Then, we fuse all embeddings (including distance, color, angle, and Gaussian embeddings, where the structure of the color distance and angle embeddings is shown in A.1) to obtain the final 3DGS embedding $E_{GSE}$. This embedding method ensures that the point cloud information in each layer is adequately represented.\\
\textbf{3DGS self-Attention.} To enhance the feature extraction from this part, we introduce the 3DGS self-attention mechanism. First, based on the 3DGS embeddings, we generate the query keys and values as follows:     
\begin{equation}
	Q_i = E^{i}_{GSE} W_{\Omega}, \quad K_j = E^{j}_{GSE} W_k + E^{j}_{GSE} W_K, \quad V_j = E^{j}_{GSE} W_v,
	\label{eq20}
\end{equation}
where $E_{GSE}^i \in \mathbb{R}^{d_t}$ represents the feature of the iii-th joint embedding, $F$ represents the corresponding feature set (as shown in the Fig.~\ref{fig2}), and $E_{Geo}^i \in \mathbb{R}^{d_t}$ represents the collection embedding, which is a combination of color-distance embedding and angle embedding (specifics can be found in A.2). $W_Q$, $W_K$, $W_V \in R^{d_t \times d_t}$are learnable projection matrices used to generate the query, key, and value. Based on this, we obtain a new attention calculation method (the specific structure shown in the Fig.~\ref{fig5} below) :
\begin{equation}
	a_{ij} = Softmax \left( \frac{Q_i K_j^T}{\sqrt{d_t}} + \lambda_g \log(1 + E^{ij}_{geo}) \right) \cdot V_j.
	\label{eq21}
\end{equation}
where, $\lambda_g \in [0.01, 1]$ controls the weight of the geometric information in the attention computation by adjusting the influence of the $\log(1 + E^{ij}_{geo})$ term. This term is designed to emphasize the relationship between geometric features, such as color and angle, thus improving the feature fusion process.
\begin{figure}[ht]
	\centering
	\includegraphics[width=0.8\textwidth]{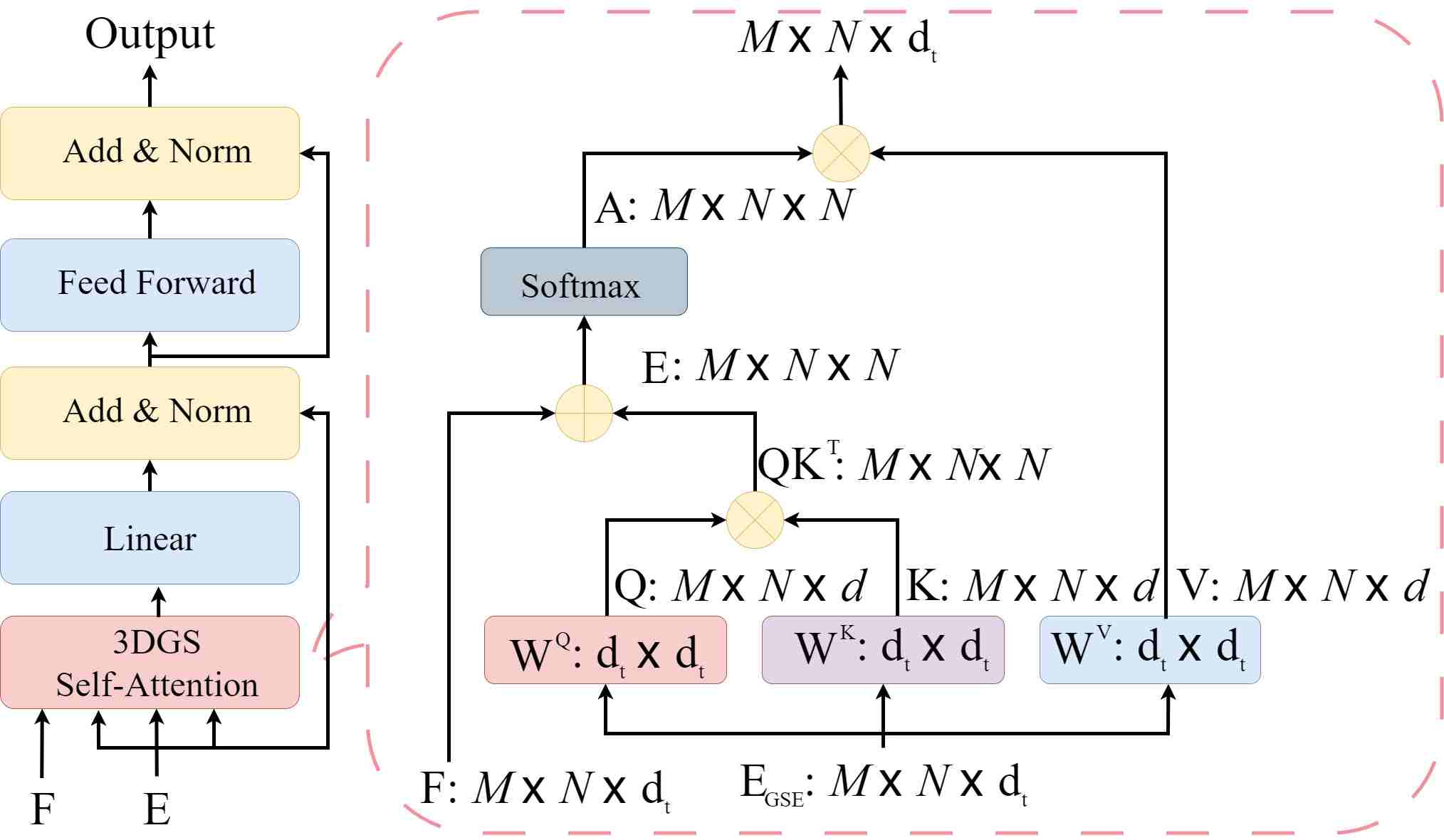}
	\caption{Left: The structure of 3DGS self-attention module. Right: The computation graph of 3DGS self-attention.}
	\label{fig5}
\end{figure}\\
\textbf{Color Distance embedding.} Colorized position embedding with color is achieved by calculating the Euclidean distance between each pair of 3DGS in the point cloud and normalizing the distance before embedding:
\begin{equation}
	d_{\text{indices}} = \frac{\| p_i - p_j \|_2}{\sigma_d},
	\label{eq22}
\end{equation}
\begin{equation}
	E_{\text{HD}} = PE(\Delta H \cdot d) W_{\text{HD}},
	\label{eq23}
\end{equation}
Where, $\sigma_d$ is the sensitivity parameter for distance control, $\Delta H =\left\| {hp}_i - {hp}_j \right\|_2$ represents the hue difference containing color information, $PE$ is the sine position embedding function, and  is the projection matrix.\\
\textbf{Angle embedding.} Angle embedding is based on the cross product of three points in the point cloud. For a given triplet of points, $p_i$, $p_j$ and their neighbor $p_k$, we calculate the angle between the three points:
\begin{equation}
	Ang_{ijk} = \max_{k \in K} \, PE\left( \frac{a \, \tan 2(\| v_i \times v_j \|, v_i \cdot v_j)}{\sigma_a} \right) W_a.
	\label{eq24}
\end{equation}
where $v_i = p_i - p_k$, $v_j = p_j - p_k$ are the vectors from $p_k$, and $\sigma_a$ is the temperature parameter for angle embedding, controlling the sensitivity of angle embedding. $W_a$is the projection matrix.
\subsection{Metrics}
\label{appendixA3}
We evaluate the performance of GeGS-PCR using standard metrics, including the Registration Recall Rate (RR), which is the proportion of point clouds with a transformation error below a threshold (e.g., RMSE < 0.2), the Feature Matching Recall Rate (FMR), representing the proportion of point cloud pairs with an inlier ratio above a threshold (e.g., 5\%), and the Inlier Ratio (IR), which measures the proportion of matching points with residuals below a threshold (e.g., 0.1m). Additionally, we assess the model's performance using the Relative Rotation Error (RRE) and Relative Translation Error (RTE). \\
\textit{Inlier Ratio (IR)} is used to measure the proportion of correspondences with geometric consistency under the true transformation. For example, given a point cloud pair $(P, Q)$ and its correspondence M, the definition is:
\begin{equation}
	IR(P, Q) = \frac{1}{|M|} \sum_{(p_i, q_j) \in M} I(\| T_{gt}(p_i) - q_j \|_2 < \delta_{corr}),
	\label{eq25}
\end{equation}
where $T_{gt}$ is the ground truth transformation matrix, and $\delta_{corr} < 0.1$ is the inlier threshold function, and $I(\cdot)$ is the Iversion bracket.\\
\textit{Feature Matching Recall (FMR)} is used to evaluate whether the correspondence geometry contains enough inliers to support registration. For N point cloud pairs, the formula is:
\begin{equation}
	FMR = \frac{1}{N} \sum_{k=1}^{N} I(IR(P_k, Q_k) \geq \eta),
	\label{eq26}
\end{equation}
where $\eta \geq 0.05$ is the minimum inlier ratio threshold required for robust registration (e.g., RANSAC).\\
\textit{Registration Recall (${RR}_{Indoor}$)} measures the proportion of successfully registered point cloud pairs that meet the geometric error tolerance. If the RMSE of the true correspondences $C^*$ of a point cloud pair $(P, Q)$ after registration satisfies the following. Then, for a threshold $\gamma \leq 0.2m$, we compute:
\begin{equation}
	RMSE = \sqrt{\frac{1}{C^*} \sum_{(p_i, q_j) \in C^*} \| T_{\text{est}}(p_i) - q_j \|_2^2},
	\label{eq27}
\end{equation}
\begin{equation}
	RR_{\text{Indoor}} = \frac{1}{N} \sum_{k=1}^{N} I(RMSE_k < \gamma).
	\label{eq28}
\end{equation}
\textit{Patch Inlier Ratio (PIR)} represents the proportion of actual overlap for superpoints (patches) under true transformations in the scene, which measures the registration of our 3DGS parameterized domain:
\begin{equation}
	PIR = \frac{1}{\hat{C}} \sum_{\substack{(sp,sq) \in C}} I \left( \exists (p_i, q_j) \in S_p \times S_q, \| T_{gt}(p_i) - q_j \|_2 < \zeta \right),
	\label{eq29}
\end{equation}
where $\zeta \leq 0.05m$ is the inlier distance threshold.\\
\textit{Pose Deviation Error} indicates the error between the estimated rigid transformation \( T_{est} = \{R, t\} \) and the true transformation \( T_{gt} = \{R_{gt}, t_{gt}\} \). We use two metrics: Relative Rotation Error (RRE) and Relative Translation Error (RTE):
\begin{equation}
	RRE = \arccos \left( \frac{{trace}(R^{T}R_{gt}) - 1}{2} \right) \quad [{radians}],
	\label{eq30}
\end{equation}
\begin{equation}
	RTE = \| t - t_{gt} \|_2 \quad [{meters}].
	\label{eq31}
\end{equation}
\textit{Registration Recall ($RR_{Outdoor}$)} measures the error in point cloud registration in outdoor scenes, considering both RRE and RTE while ensuring both satisfy the minimum threshold requirements:
\begin{equation}
	RR_{Outdoor} = \frac{1}{N} \sum_{k=1}^{N} I \left( RRE_k < 3^\circ \land RTE_k < 1.5m \right).
	\label{eq32}
\end{equation}
\subsection{Color Data}
\label{appendixA4}
\textbf{Color3DMatch/Color3DLoMatch: }Color3DMatch and Color3DLoMatch are pre-processed training datasets, which are the results of coloring the 3DMatch and 3DLoMatch point cloud data. 3DMatch contains 62 scenes, with 46 used for training, 8 for testing, and 8 for validation. The entire experimental evaluation strictly follows the dataset protocol. In 3DMatch, the overlap between point cloud pairs is greater than 30\%. In Color3DLoMatch, the overlap is relatively low (10\%-30\%).\\
\textbf{ColorKitti: }The Kitti odometry dataset contains outdoor driving scenes with 11 sequences, all obtained through LIDAR scanning. We use the corresponding RGB images for each frame to color the point clouds, thus constructing ColorKitti. However, due to the limited viewpoint of the RGB image capturing device, we only color the point clouds visible from the image perspective. Sequences 0-5 are used for training, sequences 6 and 7 for validation, and sequences 8-10 for testing. As with 3DMatch/3DLoMatch, the ground truth poses are refined using ICP, and the entire evaluation process strictly follows the dataset protocol. Additionally, only point clouds with a distance of at least 10 meters are used for evaluation. Unlike the 3DMatch/3DLoMatch scenes, the RGB images in the Kitti dataset are not captured in a panoramic manner, as the camera’s viewpoint is limited. Initially, we attempted to use neighboring frames to color the point clouds of the current frame. However, visual analysis revealed that the coloring effect was unrealistic and the color distribution was not smooth, introducing excessive color noise, which negatively impacted point cloud registration. Therefore, we only color the point clouds visible from the image perspective, and normalize the regions that are not colored. During the data preprocessing stage, we filter out colorless data and retain only the geometric point clouds. In the 3DGS registration process, only the colored portion of the point clouds is registered, and the rigid transformation parameters $(R, t)$ are applied to restore the point clouds into the global scene. Fig.~\ref{fig8} provides an example comparing the colored ground truth with the original ground truth.
\subsection{Additional Experiments} 
\label{appendixA5}
C3DM consists of 62 scenes, with 46 scenes used for training, 8 for validation, and 8 for testing. The entire process follows the C3DM/C3DLM protocol and includes registration analysis for low-overlap scenarios. Additionally, ColorKitti Odometry \cite{ao2021spinnet} consists of 11 outdoor driving sequences scanned by LiDAR. We follow the setup in \cite{qin2022geometric, yu2021cofinet}, using sequences 0-5 for training, sequences 6-7 for validation, and sequences 8-10 for testing. As in [4, 8, 16], the ground-truth poses are refined with ICP, and we only use point cloud pairs that are at least 10m apart for evaluation.\\
\textbf{Implementation details.} We implemented and evaluated our GeGS-PCR using PyTorch \cite{qin2022geometric} on an AMD 610M CPU and an NVIDIA RTX 4070 GPU. During the entire training process, GeGS-PCR was trained for 40 epochs on the 3DMatch dataset and 80 epochs on the KITTI dataset. The batch size was set to 1, with an initial learning rate of $10^{-4}$, decaying by 0.05 every epoch. The Adam optimizer was used throughout the training process.
\begin{wrapfigure}{r}{0.65\textwidth} 
	\centering
	\includegraphics[width=\linewidth]{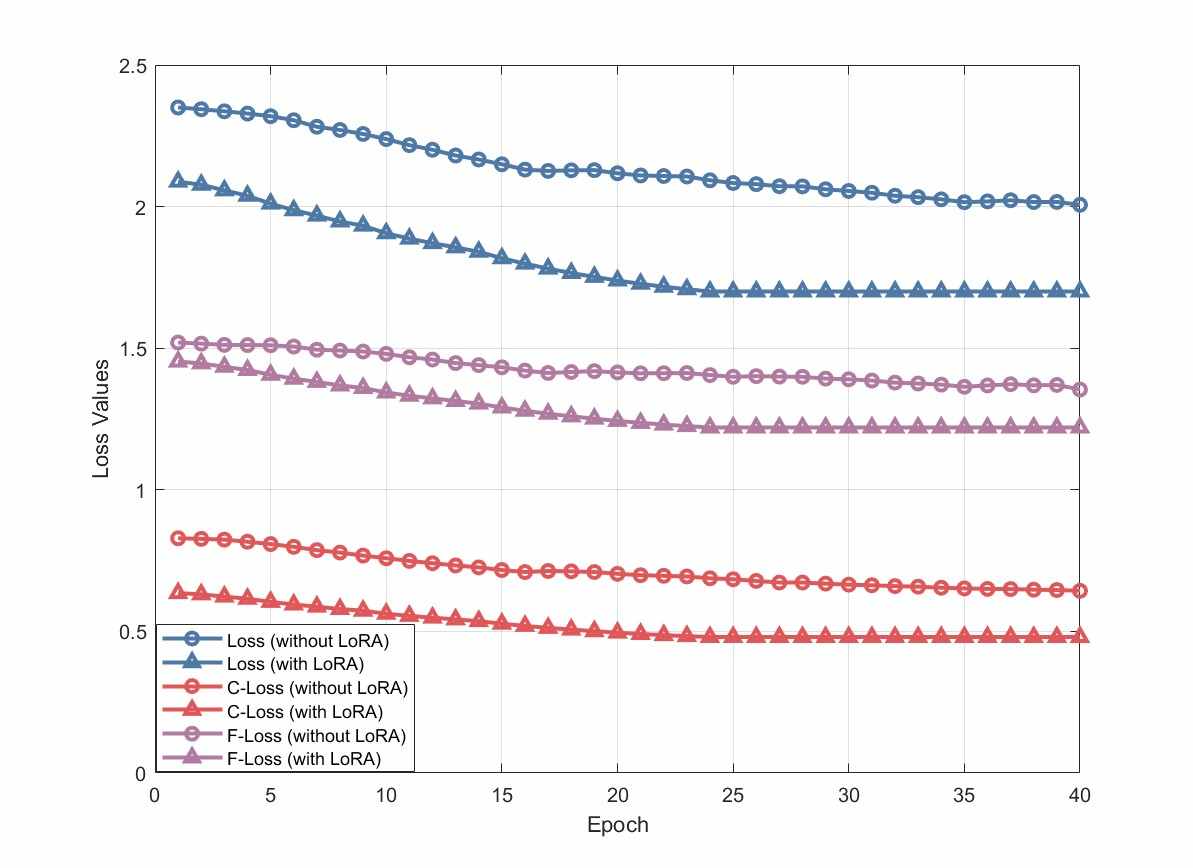}
	\caption{Comparison of Training Loss with and without LoRA Optimization (Color3DMatch Dataset).}
	\label{fig6}
\end{wrapfigure}
\textbf{Additional baselines.} Table ~\ref{tab6} is a continuation of Table ~\ref{tab1}, presenting the comparison between GeGS-PCR and other baselines. Most methods use RANSAC, and we have adopted the same approach for consistency. The results in the table show that GeGS-PCR performs excellently across all three key metrics: Feature Matching Recall (FMR), Inlier Ratio (IR), and Registration Recall (RR). In Feature Matching Re call (FMR), GeGS-PCR achieves 99.5\% on C3DM and 97.6\% on C3DLM, outperforming other methods. Particularly on C3DM, GeGS-PCR improves by 0.7\% over YOHO and significantly surpasses methods like Predator and SpinNet. In Inlier Ratio (IR), GeGS-PCR stands out further with IR values of 89.1\% on C3DM and 70.3\% on C3DLM, greatly exceeding YOHO, Predator, and other methods, especially improving by nearly 15\% on C3DM. In Registration Recall (RR), GeGS-PCR performs at 97.9\% on C3DM and 90.7\% on C3DLM, leading other methods, with a particularly large improvement of 5.2\% over YOHO on C3DLM. Overall, GeGS-PCR demonstrates outstanding performance across multiple datasets, excelling in low-overlap scenarios and surpassing all existing methods.

\begin{table}[ht]
	\caption{Evaluation results on C3DM and C3DLM. \#Samples in the table represents the number of correspondences selected by RANSAC.}
	\label{tab6}
	\centering
	\resizebox{\textwidth}{!}{  
		\begin{tabular}{l|ccccc|ccccccc}
			\toprule
			& \multicolumn{5}{c|}{C3DM} & \multicolumn{5}{c}{ C3DLM} \\
			\#Samples & 5000 & 2500 & 1000 & 500 & 250 & 5000 & 2500 & 1000 & 500 & 250 \\
			\midrule
			\multicolumn{11}{c}{Feature Matching Recall (\%) $\uparrow$} \\
			\hline
			PerfectMatch \cite{gojcic2019perfect} & 95.0 & 94.3 & 92.9 & 90.1 & 82.9 & 63.6 & 61.7 & 53.6 & 45.2 & 34.2 \\
			FCGF \cite{choy2019fully} & 97.4 & 97.3 & 97.0 & 96.7 & 96.6 & 76.6 & 75.4 & 74.2 & 71.7 & 67.3 \\
			D3Feat \cite{bai2020d3feat} & 95.6 & 95.4 & 94.5 & 94.1 & 93.1 & 67.3 & 66.7 & 67.0 & 66.7 & 66.5 \\
			SpinNet \cite{ao2021spinnet} & 97.6 & 97.2 & 96.8 & 95.5 & 94.3 & 75.3 & 74.9 & 72.5 & 70.0 & 63.6 \\
			Predator \cite{huang2021predator} & 96.6 & 96.6 & 96.5 & 96.3 & 96.5 & 78.6 & 77.4 & 76.3 & 75.7 & 75.3 \\
			YOHO \cite{wang2022you} & 98.2 & 97.6 & 97.5 & 97.7 & 96.0 & 79.4 & 78.1 & 76.3 & 73.8 & 69.1 \\
			GeGS-PCR (ours) & \textbf{99.5} & \textbf{99.6} & \textbf{99.5} & \textbf{99.7} & \textbf{99.6} & \textbf{97.6} & \textbf{97.4} & \textbf{97.1} & \textbf{97.2} & \textbf{97.0} \\
			\midrule
			\multicolumn{11}{c}{Inlier Ratio (\%) $\uparrow$} \\
			\hline
			PerfectMatch \cite{gojcic2019perfect} & 36.0 & 32.5 & 26.4 & 21.5 & 16.4 & 11.4 & 10.1 & 8.0 & 6.4 & 4.8 \\
			FCGF \cite{choy2019fully} & 56.8 & 54.1 & 48.7 & 42.5 & 34.1 & 21.4 & 20.0 & 17.2 & 14.8 & 11.6 \\
			D3Feat \cite{bai2020d3feat} & 39.0 & 38.8 & 40.4 & 41.5 & 41.8 & 13.2 & 13.1 & 14.0 & 14.6 & 15.0 \\
			SpinNet \cite{ao2021spinnet} & 47.5 & 44.7 & 39.4 & 33.9 & 27.6 & 20.5 & 19.0 & 16.3 & 13.8 & 11.1 \\
			Predator \cite{huang2021predator} & 58.0 & 58.4 & 57.1 & 54.1 & 49.3 & 26.7 & 28.1 & 28.3 & 27.5 & 25.8 \\
			YOHO \cite{wang2022you} & 64.4 & 60.7 & 55.7 & 46.4 & 41.2 & 25.9 & 23.3 & 22.6 & 18.2 & 15.0 \\
			GeGS-PCR (ours) & \textbf{76.3} & \textbf{82.4} & \textbf{86.3} & \textbf{86.6} & \textbf{89.1} & \textbf{53.4} & \textbf{58.7} & \textbf{66.9} & \textbf{69.7} & \textbf{70.3} \\
			\midrule
			\multicolumn{11}{c}{Registration Recall (\%) $\uparrow$} \\
			\hline
			PerfectMatch \cite{gojcic2019perfect} & 78.4 & 76.2 & 71.4 & 67.6 & 50.8 & 33.0 & 29.0 & 23.3 & 17.0 & 11.0 \\
			FCGF \cite{choy2019fully} & 85.1 & 84.7 & 83.3 & 81.6 & 71.4 & 40.1 & 41.7 & 38.2 & 35.4 & 26.8 \\
			D3Feat \cite{bai2020d3feat} & 81.6 & 84.5 & 83.4 & 82.4 & 77.9 & 37.2 & 42.7 & 46.9 & 43.8 & 39.1 \\
			SpinNet \cite{ao2021spinnet} & 88.6 & 86.6 & 85.5 & 83.5 & 70.2 & 59.8 & 54.9 & 48.3 & 39.8 & 26.8 \\
			Predator \cite{huang2021predator} & 89.0 & 89.9 & 90.6 & 88.5 & 86.6 & 59.8 & 61.2 & 62.4 & 60.8 & 58.1 \\
			YOHO \cite{wang2022you} & 90.8 & 90.3 & 89.1 & 88.6 & 84.5 & 65.2 & 65.5 & 63.2 & 56.5 & 48.0 \\
			GeGS-PCR (ours) & \textbf{97.9} & \textbf{97.6} & \textbf{97.5} & \textbf{96.7} & \textbf{97.6} & \textbf{90.7} & \textbf{90.2} & \textbf{90.4} & \textbf{90.0} & \textbf{89.8} \\
			\bottomrule
		\end{tabular}
	}
\end{table}
\textbf{3DGS self-attention.} In Table ~\ref{tab7}, we present the performance of different self-attention models in low-overlap scenarios, including Vanilla Self-attention, Geometric Self-attention, and 3DGS Self-attention. GeGS-PCR performs exceptionally well, particularly in low-overlap regions. Compared to other methods, 3DGS Self-attention significantly improves PIR (Precision), IR (Inlier Ratio), and RR (Registration Recall). In the high-overlap (0.9-1.0) range, the 3DGS Self-attention model achieves a PIR of 0.997, an IR of 0.905, and an RR of 1.000, demonstrating the best performance. As the overlap decreases, GeGS-PCR maintains strong performance even in the 0.5-0.6 overlap range, with a PIR of 0.938, an IR of 0.872, and an RR of 0.963, highlighting its robustness when handling low-overlap data. Specifically, compared to Vanilla Self-attention, 3DGS Self-attention shows stronger robustness across the entire overlap range, with its advantages becoming more pronounced in complex environments.

\begin{table}[ht]
	\caption{Performance of Self-attention Models}
	\label{tab7}
	\centering
	\resizebox{\textwidth}{!}{
		\begin{tabular}{l|ccc|ccc|ccc}
			\toprule
			& \multicolumn{3}{c|}{Vanilla Self-attention} & \multicolumn{3}{c|}{Geometric Self-attention} & \multicolumn{3}{c}{3DGS Self-attention}\\
			Overlap & PIR(\%) & IR(\%) & RR(\%) & PIR(\%) & IR(\%) & RR(\%) & PIR(\%) & IR(\%) & RR(\%) \\
			\midrule
			0.9-1.0 & 0.974 & 0.829 & 1.000 & 0.989 & 0.894 & 1.000 & \textbf{0.990} & 0.905 & 1.000 \\
			0.8-0.9 & 0.948 & 0.787 & 1.000 & 0.969 & 0.859 & 1.000 & \textbf{0.969} & 0.873 & 1.000 \\
			0.7-0.8 & 0.902 & 0.731 & 0.931 & 0.935 & 0.815 & 0.931 & \textbf{0.938} & 0.853 & 0.980 \\
			0.6-0.7 & 0.884 & 0.686 & 0.933 & 0.939 & 0.783 & 0.946 & \textbf{0.947} & 0.905 & 0.968 \\
			0.5-0.6 & 0.843 & 0.644 & 0.957 & 0.913 & 0.750 & 0.970 & \textbf{0.920} & 0.872 & 0.963 \\
			0.4-0.5 & 0.787 & 0.579 & 0.935 & 0.867 & 0.689 & 0.944 & \textbf{0.872} & 0.824 & 0.953 \\
			0.3-0.4 & 0.716 & 0.523 & 0.917 & 0.818 & 0.644 & 0.940 & \textbf{0.825} & 0.791 & 0.944 \\
			0.2-0.3 & 0.560 & 0.406 & 0.781 & 0.666 & 0.518 & 0.839 & \textbf{0.669} & 0.762 & 0.913 \\
			0.1-0.2 & 0.377 & 0.274 & 0.639 & 0.466 & 0.372 & 0.705 & \textbf{0.512} & 0.669 & 0.866 \\
			\bottomrule
		\end{tabular}
	}
\end{table}
\textbf{Ablation experiment.} Additionally, in Table ~\ref{tab8}, we conduct a detailed analysis of the performance of recent techniques in terms of Relative Rotation Error (RRE) (the distance between the predicted rotation matrix and the true rotation matrix) and Relative Translation Error (RTE) (the Euclidean distance between the predicted translation vector and the true translation vector). From the comparison of results, it is evident that our GeGS-PCR consistently achieves better performance.

\begin{table}[ht]
	\caption{Performance of RRE and RTE on C3DM and C3DLM.}
	\label{tab8}
	\centering
	\footnotesize
	\resizebox{\textwidth}{!}{
		\begin{tabular}{l|c|c c|c c}
			\toprule
			\multirow{2}{*}{Model} & \multirow{2}{*}{Estimator} & \multicolumn{2}{c|}{C3DM} & \multicolumn{2}{c}{C3DLM} \\
			& & RRE(°) & RTE(m) & RRE(°) & RTE(m) \\
			\midrule
			Predator \cite{huang2021predator} & RANSAC-50k & 2.029 & 0.064 & 3.048 & 0.093 \\
			CoFiNet \cite{yu2021cofinet} & RANSAC-50k & 2.002 & 0.064 & 3.271 & 0.090 \\
			GeoTransformer \cite{qin2022geometric} & RANSAC-free & 1.772 & 0.061 & 2.849 & 0.088 \\
			REGTR \cite{yew2022regtr} & RANSAC-free & 1.567 & 0.049 & 2.827 & 0.077 \\
			PEAL \cite{yu2023peal} & RANSAC-free & 1.748 & 0.062 & 2.788 & 0.087 \\
			ColorPCR \cite{mu2024colorpcr} & RANSAC-free & 1.492 & 0.048 & 2.581 & 0.075 \\
			GeGS-PCR & RANSAC-free & \textbf{1.112} & \textbf{0.024} & \textbf{2.293} & \textbf{0.051} \\
			\bottomrule
		\end{tabular}
	}
\end{table}
In Table~\ref{tab9}, we compare the performance of five loss functions in point cloud registration: (a) Cross-entropy loss, (b) Weighted cross-entropy loss, (c) Circle loss, (d) Overlap-aware circle loss, and (e) Photometric optimization loss. As the loss function improves from cross-entropy to photometric optimization, the performance on all key metrics (PIR, FMR, IR, RR) also improves. The photometric optimization loss achieves the highest performance with 87.6\% PIR, 98.2\% FMR, 71.6\% IR, and 91.9\% RR on C3DM, and 56.1\% PIR, 89.3\% FMR, 44.2\% IR, and 75.7\% RR on C3DLM, outperforming all other methods.\\
Fig.~\ref{fig6} shows the training loss curves for both the standard model (without LoRA) and the model with LoRA applied on the Color3DMatch dataset. It can be observed that for all loss types (Loss, C-Loss, and F-Loss), the model with LoRA consistently outperforms the one without LoRA, as indicated by the lower loss values throughout the training process. Specifically, the losses for the LoRA-enhanced model decrease more steadily and reach a lower final value compared to the model without LoRA, suggesting that the LoRA optimization aids in faster convergence and better performance during training. This indicates that LoRA contributes effectively to improving the model's training efficiency and overall performance in point cloud registration tasks.
\subsection{Limitations}
\label{appendixA6}
GeGs-PCR relies on superpoints (patches) extracted through downsampling during the registration process. In regions with high point cloud overlap, the large number of superpoints, combined with 3DGS parameterization, can result in significant memory usage and computational overhead. Since patch decomposition shares similarities with semantic scene understanding, we plan to leverage semantic scene understanding for point cloud pair registration in future research. Furthermore, our current registration method involves pairing after 3DGS parameterization of domain superpoints. In future work, we aim to explore scene-level registration of 3DGS for more realistic environmental registration.
\begin{table}[ht]
	\caption{Performance of ablation experiments}
	\label{tab9}
	\centering
	\resizebox{\textwidth}{!}{
		\begin{tabular}{l|cccc|cccc}
			\bottomrule
			& \multicolumn{4}{c|}{C3DM} & \multicolumn{4}{c}{C3DLM}\\ 
			Overlap & PIR(\%) & FMR(\%) & IR(\%) & RR(\%) & PIR(\%) & FMR(\%) & IR(\%) & RR(\%) \\
			\midrule
			(a) Cross-entropy loss & 80.0 & 97.7 & 65.7 & 90.0 & 45.9 & 85.1 & 37.4 & 68.4 \\
			(b) Weighted cross-entropy loss & 83.2 & \textbf{98.0} & 67.4 & 90.0 & 49.0 & 86.2 & 38.6 & 70.7 \\
			(c) Circle loss & 85.1 & 97.8 & 69.5 & 90.4 & 51.5 & 86.1 & 41.3 & 71.5 \\
			(d) Overlap-aware circle loss & 86.1 & 97.7 & 70.3 & 91.5 & 54.9 & 88.1 & 43.3 & 74.0 \\
			(e) Photometric optimization loss & \textbf{87.6} & 97.9 & \textbf{71.6} & \textbf{91.9} & \textbf{56.1} & \textbf{89.3} & \textbf{44.2} & \textbf{75.7} \\
			\bottomrule
		\end{tabular}
	}  
\end{table}
\subsection{Qualitative Results}
\label{appendixA8}
\begin{figure}[ht]
	\centering
	\includegraphics[width=\linewidth]{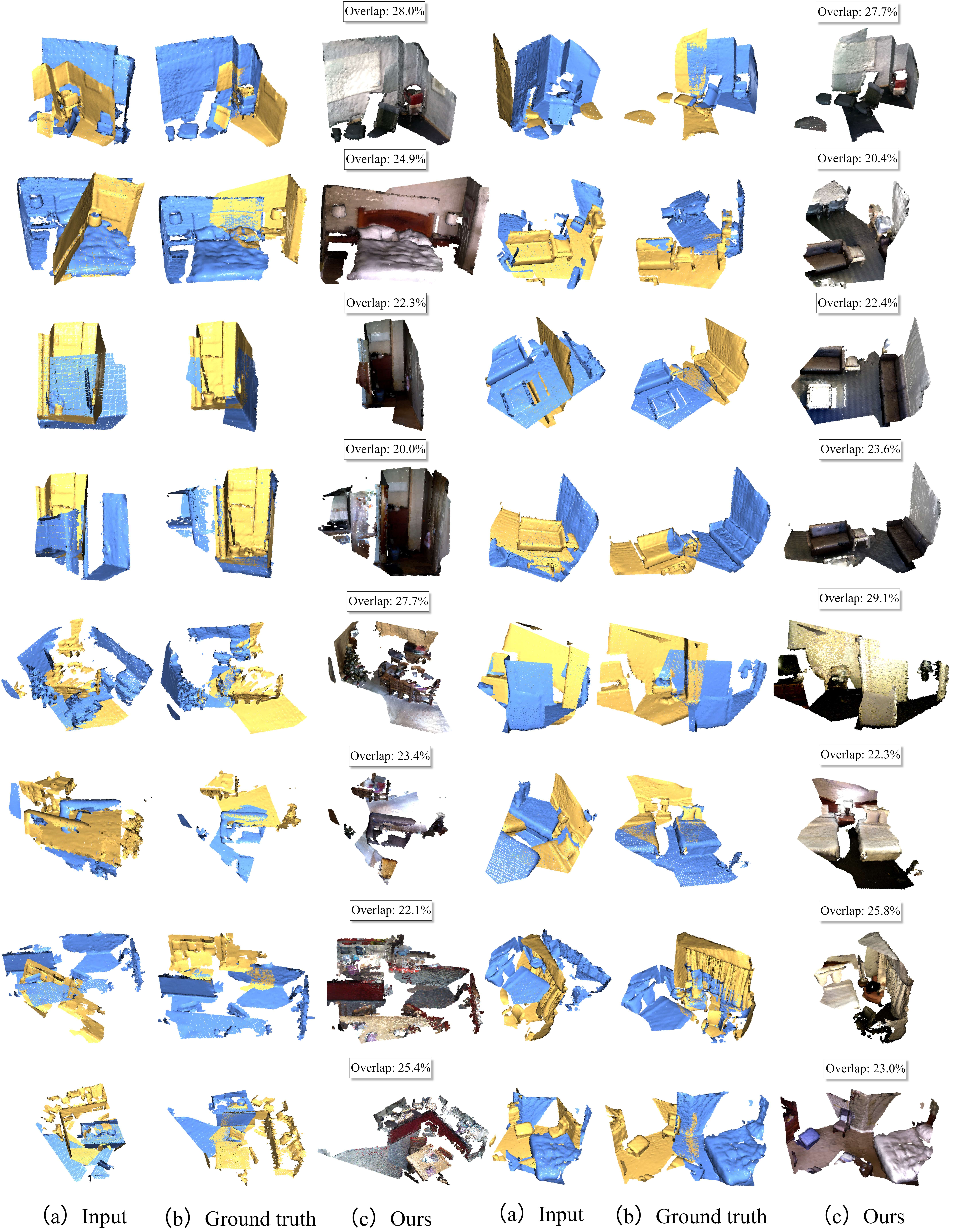}
	\caption{Registration results on Color3DMatch and Color3DLoMatch.}
	\label{fig7}
\end{figure}

\begin{figure}[ht]
	\centering
	\includegraphics[width=\linewidth]{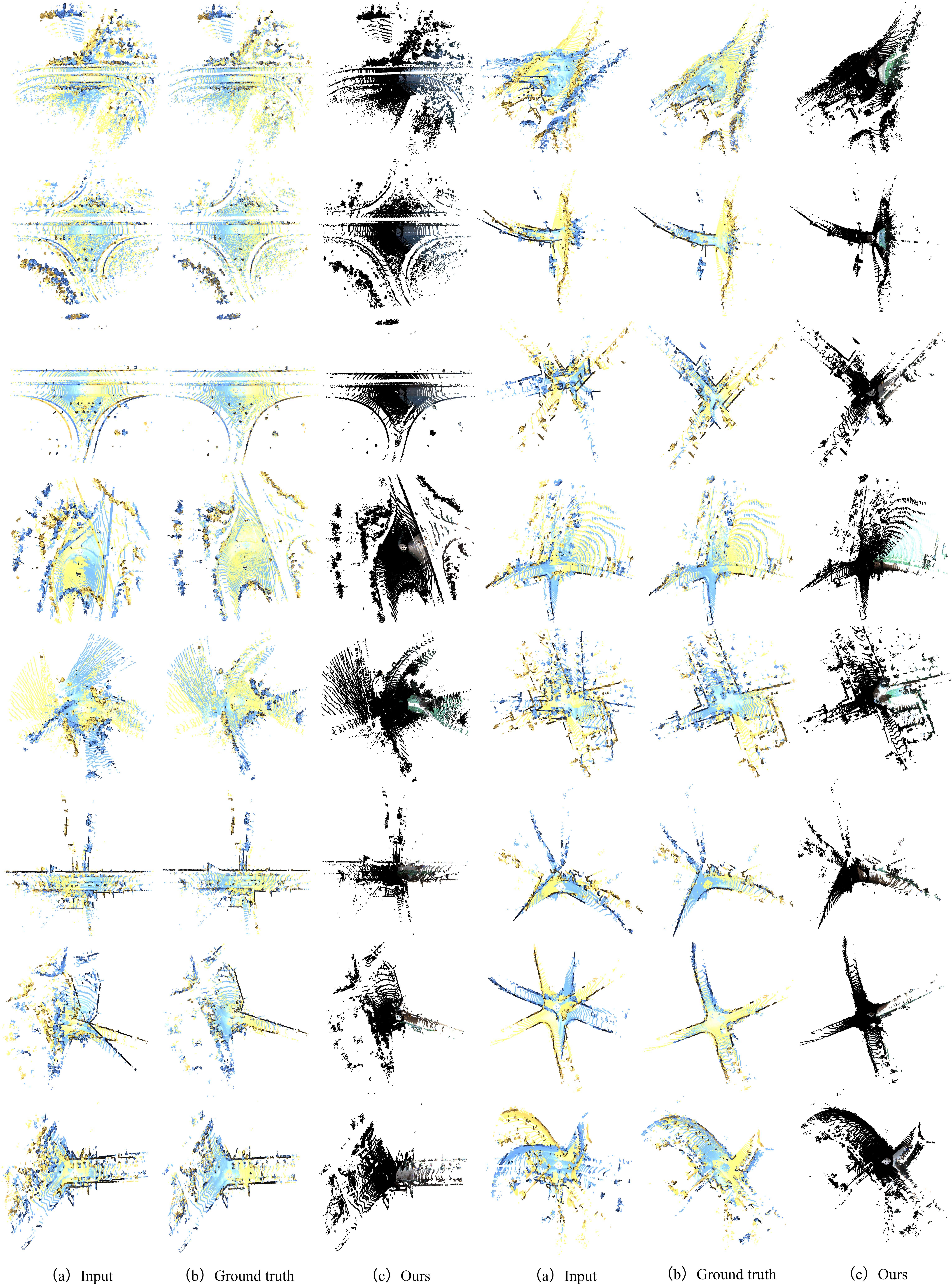}
	\caption{Registration results on ColorKitti.}
	\label{fig8}
\end{figure}

\begin{figure}[ht]
	\centering
	\includegraphics[width=\linewidth]{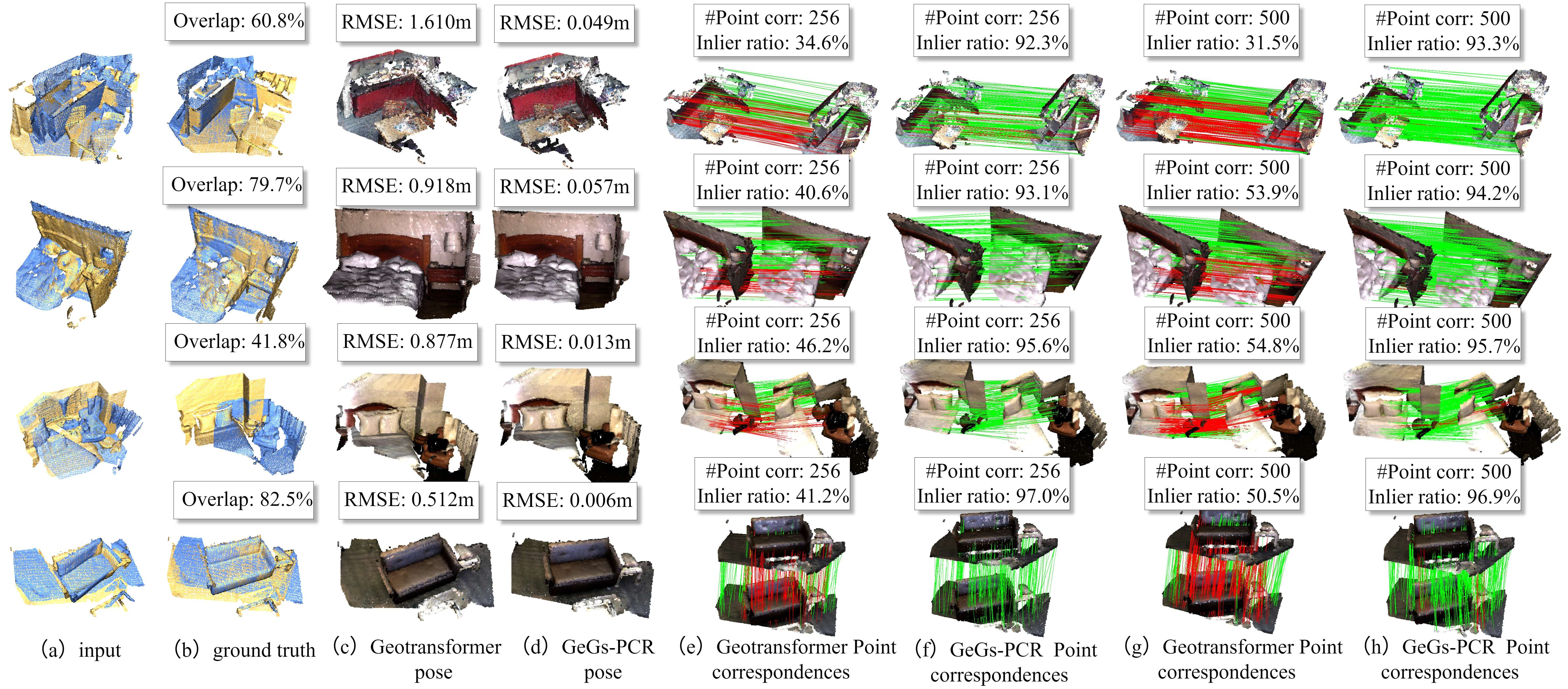}
	\caption{Registration performance with GeGS-PCR and Geometric Self-Attention across various overlap conditions.}
	\label{fig9}
\end{figure}

Fig.~\ref{fig7} and Fig.~\ref{fig8} show the registration results of GeGS-PCR on the Color3DMatch (C3DM), Color3DLoMatch (C3DLM), and ColorKitti datasets. GeGS-PCR achieves precise registration in scenarios with low overlap and subtle geometric features, demonstrating its exceptional performance in handling complex environments. From the results, it can be observed that even with smaller overlapping regions between point clouds, GeGS-PCR is still able to accurately align different point clouds. In particular, in several scenes from Fig.~\ref{fig6}, GeGS-PCR effectively handles point cloud pairs with minimal overlap and is able to precisely reconstruct the spatial structure of the point clouds. In the ColorKitti dataset shown in Fig.~\ref{fig8}, GeGS-PCR demonstrates its robustness, providing accurate registration results even in complex scenarios.\\
Fig.~\ref{fig9} shows the registration performance comparison between GeGS-PCR and Geotransformer across various overlap conditions. The results demonstrate that in high overlap scenarios, GeGS-PCR exhibits superior registration accuracy compared to Geotransformer, particularly in terms of point correspondence and inlier ratio. For example, with a 79.7\% overlap, GeGS-PCR achieves an inlier ratio of 93.1\%, while Geotransformer only reaches 40.6\%. In low overlap conditions, GeGS-PCR's advantage becomes even more evident. For instance, with a 41.8\% overlap, GeGS-PCR's RMSE is 0.006m, much lower than Geotransformer's 0.877m. GeGS-PCR is able to effectively capture the correct correspondences between point clouds in low-overlap scenarios, maintaining high registration accuracy. Overall, GeGS-PCR demonstrates higher stability and precision across various overlap conditions, proving its robustness in complex scenarios.\\
\textbf{Advantages. }The advantages of GeGS-PCR lie in the collaborative optimization of global and local structures. Through local Gaussian feature extraction, GeGS-PCR effectively suppresses noise interference and robustly fuses geometric and color features. In low-overlap or noisy point cloud data, GeGS-PCR dynamically adjusts local geometric distribution through covariance modeling, significantly improving registration accuracy. Additionally, photometric loss based on differentiable rendering optimizes global pose consistency, ensuring stable registration. GeGS-PCR is not only effective for high-overlap registration but also offers an efficient, scalable solution for low-overlap tasks like autonomous driving and large-scale scene reconstruction.\\
Additionally, by leveraging both geometric and color information, GeGS-PCR is able to find consistent feature correspondences in low-overlap regions, improving registration accuracy. The changes between point clouds before and after registration in the qualitative results clearly illustrate how our model handles challenging regions. Notably, in areas with similar geometric features such as floors and walls or appliances and furniture, GeGS-PCR maintains high accuracy. Overall, GeGS-PCR showcases superior registration capability and robustness when dealing with low-overlap scenarios, complex structures, and environments with rich color information.
\end{document}